\documentclass[10pt,twocolumn,letterpaper]{article}

\usepackage[table,dvipsnames]{xcolor}

\usepackage{cvpr}

\usepackage{graphicx}
\usepackage{amsmath,amssymb,amsthm}
\usepackage{mathtools}
\usepackage{booktabs}
\usepackage{comment}
\usepackage{color}
\usepackage[nice]{nicefrac}
\usepackage{multirow}

\graphicspath{{figures/}}

\usepackage{algorithm}
\usepackage[noend]{algpseudocode}
\usepackage{bm}
\usepackage{hyperref}
\usepackage{amsfonts}
\usepackage{longtable}
\usepackage{array}

\title{Domain Adaptation via Feature Refinement}
\author{\normalsize\bf Savvas Karatsiolis$^{1}$\thanks{\texttt{s.karatsiolis@cyens.org.cy}} \quad\bf Andreas Kamilaris$^{1,2}$ \\
	\normalsize$^{1}$CYENS Center of Excellence, Nicosia, Cyprus \\
	\normalsize$^{2}$University of Twente, Enschede, The Netherlands}

\begin{document}
\maketitle
\begin{abstract}
We propose Domain Adaptation via Feature Refinement ($DAFR^2$), a simple yet effective framework for unsupervised domain adaptation under distribution shift. The proposed method synergistically combines three key components: adaptation of Batch Normalization statistics using unlabeled target data, feature distillation from a source-trained model and hypothesis transfer.  By aligning feature distributions at the statistical and representational levels, $DAFR^2$ produces robust and domain-invariant feature spaces that generalize across similar domains without requiring target labels, complex architectures or sophisticated training objectives. Extensive experiments on benchmark datasets, including CIFAR10-C, CIFAR100-C, MNIST-C and PatchCamelyon-C, demonstrate that the proposed algorithm outperforms prior methods in robustness to corruption. Theoretical and empirical analyses further reveal that our method achieves improved feature alignment, increased mutual information between the domains and reduced sensitivity to input perturbations.
\end{abstract}

\section{Introduction}
Despite the remarkable results of Deep Learning (DL) models on benchmark datasets, they systematically show significant performance drops when encountering even minor distribution shifts or natural corruptions during testing \cite{1.ServeyOOD,2.NaturalAdversarialExamples,3.Robustness2DS,4.BenchmarkingNNRobustness2CorruptionsPerturbations}. Thus, the ability of Artificial Intelligence (AI) to operate reliably in dynamic environments hinges on its difficulty in processing and acting upon data that deviates from the controlled training conditions. This behavior poses a barrier to deploying AI models in the real world, especially in safety-critical domains like autonomous car driving and medical diagnosis. Unfortunately, unexpected variations in production environments are the norm, not the exception. Bridging the gap between development-time performance and real-world resilience is a fundamental challenge requiring novel approaches to advancing model generalization and adaptation. Nevertheless, there always exists irreducible uncertainty which arises from the natural complexity of the data or abnormal outliers.

Existing techniques such as Deep Ensembles \cite{5.CanYouTrustModelUncertainty, 6.DeepEnsembles2}, Monte Carlo (MC) Dropout \cite{7.MC_Dropout1,8.MC_Dropout2} and temperature scaling \cite{11.TemperatureScaling3,9.TemperatureScaling1,10.TemperatureScaling2} offer only marginal improvements in mitigating the problem. Model calibration \cite{9.TemperatureScaling1,13.ModelCalibration2,14.ModelCalibration3}, in particular, fails to transfer effectively from the source to the target domain. While Deep Ensembles generally outperform other methods, their performance remains significantly below satisfactory levels \cite{5.CanYouTrustModelUncertainty}. There is also often a trade-off between clean accuracy and robustness: many techniques enhance one while diminishing the other. The robustness and transferability of models, specifically Convolutional Neural Networks (CNNs), tend to be linked to scaling: increasing the training dataset size and the model’s capacity enhances the CNN’s robustness to distributional shifts, i.e., larger models trained on more extensive datasets demonstrate improved generalization to Out-Of-Distribution (OOD) data \cite{15.ScalingNNsAndRobustness1, 16.ScalingNNsAndRobustness2,17.ScalingNNsAndRobustness3}. Another technique for mitigating certain robustness issues involves simple modifications to data preprocessing, such as altering image resolution \cite{15.ScalingNNsAndRobustness1,18.ImageResolutionAndRobustness} or applying certain types of data augmentation \cite{19.AugmentationsAndRobustness1,  20.AugmentationsAndRobustness2, 21.AugmentationsAndRobustness3}. 

More recently, the vulnerability of deep learning models to distribution shifts in vision has been addressed through a plethora of research strategies. These endeavors seek to align data from a deviating domain more closely with the feature space of the source domain. Such methods encompass strategically selected augmentations during training \cite{22.Augmix, 22b.SimpleWay2MakeNNsRobust}, adversarial domain adaptation techniques \cite{23.AdversarialDomainAdaptation1, 24.AdversarialDomainAdaptation2, 25.AdversarialDomainAdaptation3}, self-supervised learning approaches \cite{26.SSforDA1, 27.SSforDA2}, adaptation of test domain statistics \cite{28.Batch-InstanceNormalization, 29.CovariateShiftAdaptation}, and pseudo-labeling \cite{30.Pseudo-labelsForDA1, 31.Pseudo-labelsForDA2}. A more recent and significant development is Test-Time Adaptation (TTA) \cite{32.SurveyTTA}. TTA methodologies are generally classified into two groups depending on the accessibility of the original training data during adaptation. One group comprises training data-free TTA, performing adaptation using only the test data. The other group, termed Test Time Training (TTT), integrates the training data (or source domain information) alongside the test data to direct the adaptation process. While offering substantial performance gains, many of these approaches often suffer from sensitivity to misleading patterns and hyperparameter tuning and exhibit training instability \cite{33.ContextGuidedEM}.

In this paper,  we propose a domain adaptation method based on synergistically training two models: one on labeled source data and another on unlabeled target data. This joint training paradigm enables effective information transfer from the source to the target model, resulting in excellent performance on the target domain without compromising source domain performance. By employing key ML principles, we facilitate knowledge transfer from the source to the target domain model, leading to state-of-the-art performance and the establishment of new domain adaptation benchmarks for popular datasets. Leveraging its simplicity, ease of implementation, inherent training stability, and the absence of special hyperparameters, the proposed approach achieves state-of-the-art results, establishing a new standard in the field of domain adaptation.

\section{Preliminaries}
This section discusses the key elements of our domain adaptation approach and analyzes their contribution to the task. The authors deemed it important to include these preliminary concepts to facilitate readers' comprehension and orientation before introducing the main methodology.

\subsection{Definitions}
To ensure clarity and consistency throughout this paper, we define the key symbols and notations used. To address the challenge of domain shift, our objective is to learn a robust classification model for the target domain. We assume access to a source domain $p_s$ with data instances $x \in X$ and their corresponding labels $y \in Y$. For the target domain $p_t$, we have a set of unlabeled data instances $z \in Z$. The primary task is to develop a model $\mathcal{F}_t : X \rightarrow Y$ and $\mathcal{F}_t : Z \rightarrow Y$ whose classification performance on the source data is comparable to that of a model $\mathcal{F}_s: X \rightarrow Y$ trained directly on the labeled source data. Crucially, $\mathcal{F}_t$ must be robust to the distribution shift between the source domain and the target domain, i.e., $\mathcal{F}_t : Z \rightarrow Y$ significantly outperforms $\mathcal{F}_s : Z \rightarrow Y$ when inferring on data from the target domain.

\subsection{Target domain statistics adaptation}
\paragraph{}
Batch Normalization (BN) \cite{34.BatchNormalization} is a widely used technique in training CNNs that significantly improves performance and training stability. While using BN offers several benefits during model training, its most significant impact lies in mitigating internal covariate shift—a phenomenon where changes in the parameters (weights and biases) of preceding layers alter the distribution of inputs to subsequent layers. BN mitigates this problem by normalizing the output of the model’s layers by subtracting the batch mean $\mu$ and dividing by the batch standard deviation $\sigma$. This ensures that the input to each layer has a more stable distribution (closer to zero mean and unit variance) throughout the training process. A direct consequence of stabilizing the input distributions of subsequent model layers is achieving training acceleration and faster convergence via smoother optimization landscapes \cite{34.BatchNormalization}.
\par
Another important and often overlooked property of BN is that it whitens activations by aligning them with the Principal Components (PC) of the layer’s representation, resulting in an isotropic variance. This has significant implications, including reducing feature-wise correlations and pushing the covariance matrix $\Sigma$ closer to a diagonal form $\mathbb{I}$. Throughout the training on the labeled source domain data, the network learns features aligned with the dominant eigenvectors of the activations’ matrix. 
\noindent
Given a batch of activations $X \in \mathbb{R}^{m\times d}$, where $m$ is the batch size and $d$ is the number of features, the sample covariance matrix before BN is: \\
\[ 
\Sigma \; = \; \frac{1}{m}X^TX .
\]
Expressing $\Sigma$ in its eigenvalue decomposition:
\[ 
\Sigma \; = \; V\Lambda V^T ,
\]
where $V \in \mathbb{R}^{d\times d}$ is the orthonormal matrix of eigenvectors and $\Lambda \; = \; {\bf diag}(\lambda_1, ... , \lambda_d)$ is the diagonal matrix of eigenvalues representing variances along the principal components.
\\
The covariance matrix after BN is applied becomes: 
\[
\Sigma_{\text{BN}} = \frac{1}{m} \, \hat{X}^\top \hat{X},
\]
where \( \hat{X} = D^{-\frac{1}{2}} (X - \mu) \), and 
\( D = \operatorname{diag}(\sigma_1^2, \dots, \sigma_d^2) \) 
is the diagonal matrix of feature-wise variances. Substituting $\hat{X}$, the covariance matrix becomes: 
\[
\Sigma_{\text{BN}} 
= \frac{1}{m} \left( D^{-\frac{1}{2}} (X - \mu) \right)^{\!T} 
\left( D^{-\frac{1}{2}} (X - \mu) \right).
\]
Since $D^{-\frac{1}{2}}$ is diagonal, it commutes in the product:
\[
\Sigma_{\text{BN}} = D^{-\frac{1}{2}} 
\left( \frac{1}{m} (X - \mu)^\top (X - \mu) \right) D^{-\frac{1}{2}}.
\]
The inner term is the original covariance $\Sigma$ (before BN is applied):
\[
\Sigma_{BN} \;=\; D^{-\frac{1}{2}} \; \Sigma \; D^{-\frac{1}{2}}.
\]
Substituting $\Sigma \; = \; V\Lambda V^T $ and $D^{-\frac{1}{2}} \; \approx \; \Lambda^{-\frac{1}{2}} $ 
\[
\Sigma_{BN} \;=\; \Lambda^{-\frac{1}{2}} \; ( V\Lambda V^T) \; \Lambda^{-\frac{1}{2}}.
\]
Using the fact that $\Lambda$ is diagonal and so it commutes the product: 

\[
\Sigma_{\text{BN}} = V \left( \Lambda^{-\frac{1}{2}} \Lambda \, \Lambda^{-\frac{1}{2}} \right) V^\top.
\]
Substituting $\Lambda^{-\frac{1}{2}}\Lambda\Lambda^{-\frac{1}{2}} \;= \;\mathbb{I} $, we get: 
\[
\Sigma_{BN} \;=\; V \mathbb{I} V^T .
\]
Finally, since $VV^T = \mathbb{I}$ : 
\[
\Sigma_{BN} \;=\; \mathbb{I} .
\]

We exploit this property by adjusting the BN statistics using samples from the target domain during the training of the source domain model so that the normalization process becomes more attuned to the characteristics of the target distribution. This direct adaptation of the normalization parameters helps align the feature distributions between the source and target domains at a fundamental level, i.e., learn the principal components of both domains rather than only the source domain’s components. The activations produced by the target domain are normalized based on their statistical properties, rather than the potentially misleading statistics (from the target domain's perspective) of the source domain. Accurate normalization is critical for producing discriminative and domain-invariant features. When BN statistics are adapted to the target domain, the resulting feature representations are more robust to domain shift because they are built around the principal components identified in both domains (source and target). This allows subsequent layers in the network to produce features more comparable to those seen during training (or features from an ideally aligned domain), facilitating better generalization. While numerous domain adaptation strategies achieve domain-invariant features by tuning the BN layers' parameters based on target domain data \cite{28.Batch-InstanceNormalization, 35.TheNormMustGoOn, 36.ImprovingRobustnessByCovariateShiftAdapt, 37.TowardsStableTTAinDynamicWorld,38.EffectiveRestorationOfSourceKnowledgeTTA}, this is the first work that provides a theoretical insight into how this is achieved by BN adaptation.   
\par
BN tuning leverages target domain statistics and can be performed during the training or testing phase. 
BN parameters’ adaptation using target domain samples forms a cornerstone of our domain adaptation technique. It operates synergistically with feature distillation and knowledge sharing to facilitate knowledge transfer from the source to the target domain model.

\subsection{Feature distillation}
\par
Situated within the broader framework of knowledge distillation, feature distillation involves training a model commonly referred to as the student or target model to replicate the internal representations of another model, known as the teacher or source model \cite{108.HintonDistillation}. By aligning these internal representations, the target model can learn more informative and robust features.  Notably, several studies have reported that despite having significantly smaller capacity, the student model can outperform the teacher in terms of accuracy \cite{39:FitNets, 40.ChannelDistillation, 41.StudentTeacherDeviations}. Most of these approaches rely on Mean Squared Error (MSE) loss to transfer the teacher’s feature representations. In general, knowledge distillation encompasses methods in which the student is trained to mimic the intermediate computations of a teacher model. These computations do not necessarily involve feature maps; they may instead include attention maps \cite{42.PayingAttention2Attention} or other auxiliary computations applied to the model’s processing stream \cite{43.AGiftFromKnowledgeDistillation}. Alternatively, the target model may learn to mimic the activation statistics of the source model, treating knowledge transfer as a distribution matching problem rather than direct feature matching \cite{44.NeuronSelectivityTransfer}. Similarly, knowledge transfer can be achieved by maximizing the mutual information between the source and target models using contrastive loss \cite{45.ContrastiveRepresentationDistillation} or Maximum Mean Discrepancy (MMD) between the source and target features \cite{44.NeuronSelectivityTransfer}. Notably, MSE between two fixed vectors is the special case of MMD using a linear kernel between two Dirac distributions \cite{46.KernelMeanEmbeddingOfDistributions, 47.KernelTwoSampleTest}. Despite the different strategies, all knowledge transfer techniques achieve knowledge transfer by aligning the target model’s deep representations with the source model’s representations.
\par
We use the MSE regression loss to achieve feature distillation due to its de facto ability to filter out redundant or noisy information from the teacher’s knowledge, thereby helping to retain only the most relevant aspects of the task. Specifically, we train the target model to mimic the activations of the feature extractor in the source model, that is, the output preceding the fully connected classifier, using data from both the source and target domains. This process, combined with the adaptation of target domain statistics on the source model, enables the target model to distill the core, domain-invariant features shared across the two domains. At the same time, it helps suppress domain-specific noise and idiosyncrasies arising from inter-domain discrepancies. By distilling knowledge from these refined intermediate representations, the target model learns to recover robust signals, ultimately improving generalization through the suppression of domain noise.

Given a classifier that maps the input space $X$ to an intermediate layer such as $\mathcal{F}_s: X \rightarrow E_s$, with $E_s$ being the embedding space of  $\mathcal{F}_s$: 
\[
E_s = f_s(x)\;=\; h(x) + \delta(x), 
\]
where $h(x)$ represents the robust domain features and $\delta(x)$ represents noise, artifacts, or data peculiarities that are not shared between the source and the target data domains. Our task is to align the embeddings space $E_t$ of the final model $\mathcal{F}_t$ such as:
\[
E_t = E_s = f_t(x)\;=\; h(x) + \delta(x), 
\]
Let $x\sim p_s $ denote samples from the source domain and $x\sim p_t$ denote samples from a target domain. We assume that for samples $x\sim p_s $, $\delta(x)$  is relatively small since the robust features $h(x)$ are learned by training a classifier $\mathcal{F}_s$ with samples exclusively from $p_s$ while for samples $x\sim p_t $, $\delta(x)$ is larger and the classifier behaves unpredictably on such samples because of the large $\delta(x)$ component. The regression model $\mathcal{F}_t:X \rightarrow E_t$ is trained to mimic the embeddings of the classifier $\mathcal{F}_s$ and uses MSE as a loss function: 
\[
\mathcal{L} (\mathcal{F}_t) \;=\; E_{x\sim p_s \cup p_t} \;  \lbrack \; \Vert f_t(x) - f_s(x) \Vert^2 \; \rbrack .
\]
By decomposing the error:
\begin{align*}
	\mathbb{E} \left[ ( f_t(x) - f_s(x) )^2 \right] 
	&= \mathbb{E} \left[ \left( f_t(x) - \mathbb{E} \left[ f_s(x) \mid x \right] \right. \right. \\
	&\quad \left. \left. + \mathbb{E} \left[ f_s(x) \mid x \right] - f_s(x) \right)^2 \right] =\
\end{align*}
\begin{flushright}
	\begin{align*}
		& \mathbb{E} \left[ \left( f_t(x) - \mathbb{E} \left[ f_s(x) \mid x \right] \right)^2 \right] \\
		& + 2 \mathbb{E} \left[ 
		\left( f_t(x) - \mathbb{E} \left[ f_s(x) \mid x \right] \right)
		\left( \mathbb{E} \left[ f_s(x) \mid x \right] - f_s(x) \right)
		\right] \\
		& + \mathbb{E} \left[ 
		\left( \mathbb{E} \left[ f_s(x) \mid x \right] - f_s(x) \right)^2 
		\right].
	\end{align*}
\end{flushright}
The middle term is zero since $f_t(x) - \mathbb{E} \lbrack f_s(x) \vert x \rbrack$ is a function of $x$ and by the definition of conditional expectation 
\[
\mathbb{E}  \;\lbrack \; \mathbb{E} \; \lbrack \; f_s(x)  \vert x \; \rbrack - f_s(x) \vert x \; \rbrack = 0 \ \ \Rightarrow
\]
\[
\mathbb{E} \; \lbrack \; (f_t(x) - \mathbb{E} \;\lbrack f_s(x) \; \vert  x \; \rbrack \;)\; (\mathbb{E}\; \lbrack f_s(x) \; \vert \; x \rbrack - f_s(x)) \; \rbrack \; = \; 0
\]
Thus, we have: 
\[
\mathbb{E} \; \lbrack \; ( f_t(x) - f_s(x) ) ^2 \; \rbrack  = \mathbb{E} \; \lbrack \; (f_t(x) - \mathbb{E} \lbrack f_s(x) \; \vert \; x \rbrack)^2 \rbrack + a
\]
where $a$ is some constant. The expression is minimized (pointwise in $x$) when 
\[
f_t^*(x) \;=\; \mathbb{E} \; \lbrack \; f_s(x) \vert x \; \rbrack  
\]
Substituting the decomposition $\mathcal{F}_t(x)\;=\; h(x) + \delta(x)$ into the conditional expectation:
\[
f_t^*(x) \;=\; \mathbb{E} \; \lbrack \; f_s(x) \; \vert \; x \; \rbrack \;=\; \mathbb{E} \; \lbrack \; h(x) + \delta(x) \; \vert \; x \;\rbrack .
\]
Assumming that $h(x)$ is deterministic and $\mathbb{E} \; \lbrack \; \delta(x) \; \vert \; x \; \rbrack =\;0$ we get:
\[
f_t^*(x) \;=\; h(x) + \mathbb{E} \; \lbrack \; \delta(x) \; \vert \; x \; \rbrack \;=\; h(x) + 0 = h(x).
\]

This result demonstrates that an optimal regression model recovers the robust signal $h(x)$ by eliminating the influence of the noisy component $\delta(x)$. However, while this result provides a theoretical insight into the potential of feature distillation, achieving this in practice is challenging. When using plain regression, it struggles to make $\delta(x)$ negligible, primarily due to high correlations among the source model's features, the model's unpredictable behavior to out-of-distribution data and its inability to handle unseen or noisy data. These factors result in a significant conditional expectation of the noise ($\mathbb{E} \; \lbrack \; \delta(x) \; \vert \; x \; \rbrack \neq 0 $), which prevents the regression from approaching optimality. In contrast, our proposed method combines feature distillation with target domain statistics adaptation. As shown theoretically in the previous section, this approach de-correlates the components of the two domains. This effectively nullifies the conditional expectation of the noise (($\mathbb{E} \; \lbrack \; \delta(x) \; \vert \; x \; \rbrack \approx 0 $)), thereby enabling the feature distillation to converge to a near-optimal regression model.

\subsection{Hypothesis Transfer}
\par
Since the target domain is unlabeled, we cannot directly infer decision boundaries from its data. Instead, we borrow the classifier learned on the labeled source domain whose decision boundaries are already encoded in the source model’s classifier $g_s$. By fixing $g_s$ and placing it atop our target-domain feature extractor $f_t$, we preserve those boundaries during adaptation, providing a stable objective that simplifies learning. The adaptation then reduces to teaching $f_t$ to produce features that align with the source features in a shared space. Concretely, let the source model be $\mathcal{F}_s = g_s(f_s(x))$, where $f_s$ is the source feature extractor and $g_s$ the classifier. For the target domain, we train $\mathcal{F}_t = g_s(f_t(x))$, driven by target-domain statistics alignment and feature-level distillation from $f_s$ to $f_t$. During inference, we simply feed the target sample through $h_t$ and then through the fixed $g_s$. Because $g_s$ already encodes robust class boundaries, its reuse ensures our adapted model maintains the original class structure without requiring further adaptation on the source data. Effectively, this strategy transfers the classification hypothesis inferred from the labeled source domain to the unlabeled target domain \cite{47.DoWeReallyNeed2AccessSourceData}. In the ideal case—when $f_t$ perfectly aligns the learned features with the source feature manifold—the performance of $f_t$ on the source domain approaches that of $f_s$, effectively transferring the learned classification hypothesis without requiring any labeled target data.
\par
The strategies introduced in this section form the backbone of our domain adaptation approach, enabling it to achieve state-of-the-art performance on widely used benchmark datasets. While each of the three strategies is crucial, they all center around a unifying principle: feature alignment. Adapting target domain statistics plays a key role in refining the source model’s feature learning, ensuring that the distillation process captures and transfers the aligned core components across domains. Meanwhile, hypothesis transfer is indispensable in the absence of target labels, leveraging the shared feature space learned by the target model to guide accurate predictions. Since feature alignment in our method is achieved through two complementary processes—primary component refinement (via target domain statistics adaptation) and representation refinement (via feature distillation)—we refer to our approach as Domain Adaptation via Feature Refinement ($DAFR^2$).

\subsection{Integration-Synergy}
$DAFR^2$ harnesses the three key components that work in concert to achieve effective domain adaptation. Target domain statistics adaptation focuses on aligning the activation patterns of the source model with the fundamental data structures intrinsic to both the source and target domains. This process facilitates the model's comprehension of the underlying patterns within the novel, target data. After this initial alignment, feature distillation refines the extracted features by enabling the target model to discard signals attributable to domain-specific peculiarities, such as stylistic variations, characteristics of data acquisition, or domain-inherent deformations. Feature space refinement mitigates the influence of irrelevant, domain-specific noise. Building upon these refined and aligned features, hypothesis transfer then applies the learned classification knowledge, leveraging the improved feature representation to accurately classify data within the target domain.

Collectively, target domain statistics adaptation and feature distillation operate as an effective feature refinement mechanism that increases consistency across domains rather than a transfer learning mechanism. The theoretical perspective taken in Section 2 concerning these two mechanisms justifies the powerful impact they have on the target model's feature space. Furthermore, hypothesis transfer capitalizes on this refined feature space to enable accurate classification, effectively transferring the capacity to classify features that have undergone appropriate alignment.

In the following Section, we provide a detailed explanation of how these strategies are adopted to form the $DAFR^2$ framework.

\section{Domain Adaptation via Feature Refinement}
\begin{figure*}[htbp]
	\centering
	\includegraphics[width=0.9\textwidth]{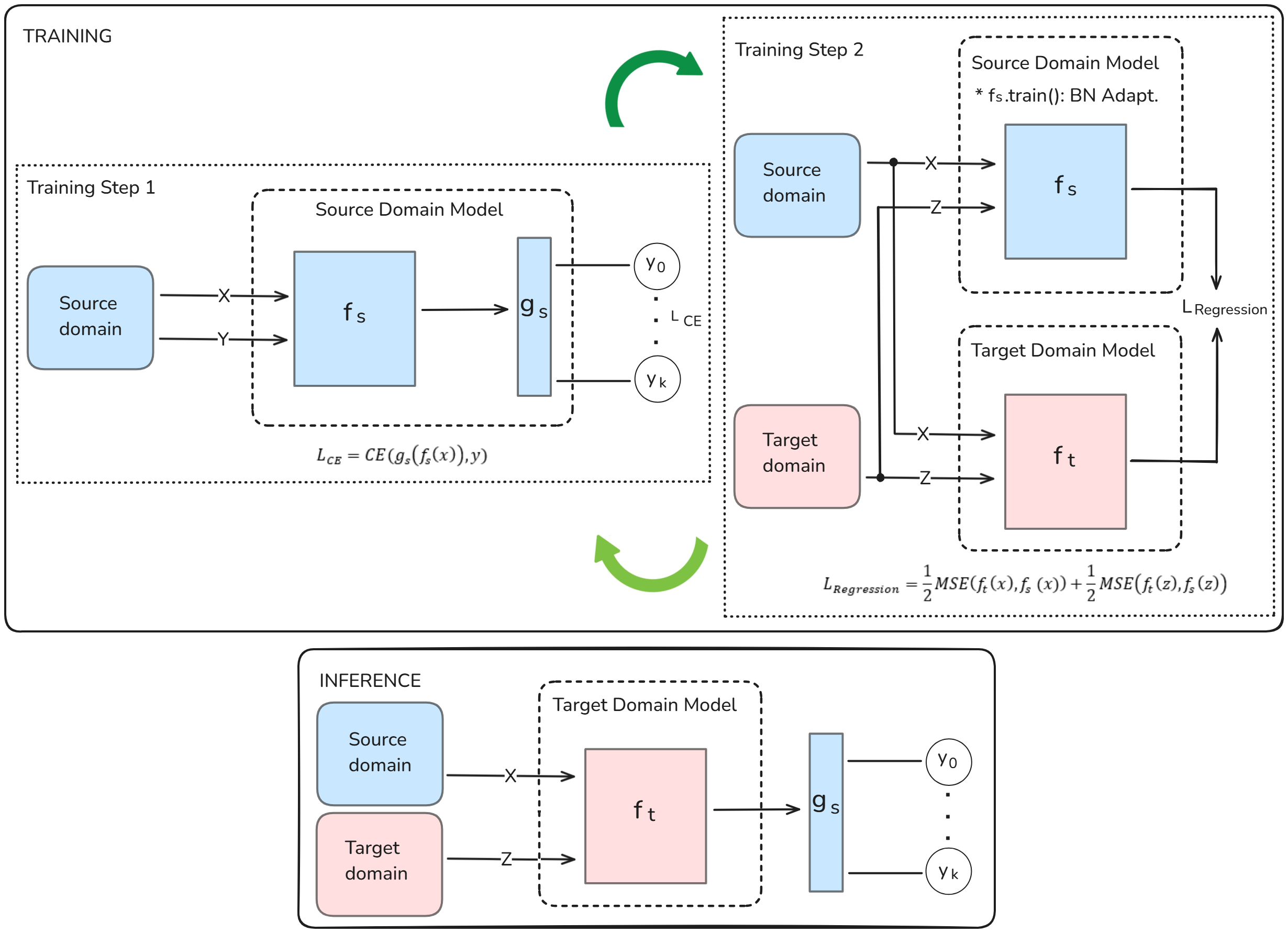}
	\caption{The $DAFR^2$ framework during the training and inference phases. Training phase comprises two steps: a) Learning the features of the source domain via supervised learning and b) Transferring this learning to the target domain model with feature distillation. During Step 2, the source model adapts its BN layers on both domains' data. At inference, both source and target domain data undergo a two-stage process: first, they are passed through the target domain model $f_t$, and subsequently, classified by the source domain classifier $g_s$.}
	\label{fig:dafr_architecture}
\end{figure*}

Our approach relies on synergistically training two models: a source domain model and a target domain model. The source model is trained using labeled source domain data to classify input samples and serves as a knowledge distillation proxy for the target model. The target model, in contrast, is trained on unlabeled data from both domains and learns to approximate the internal feature representations of the source model. While the two models do not need to be architecturally identical, the source model must include BN layers, which are crucial for adapting to target domain statistics. Figure \ref{fig:dafr_architecture} illustrates the overall $DAFR^2$ workflow.
Each training loop iteration consists of two sequential steps:
\begin{enumerate}
	\item Supervised training of the source model: The source model is updated using cross-entropy (CE) loss on labeled source domain data to improve its classification accuracy.
	\item Unsupervised training of the target model via feature distillation:
	Both source and target domain data are passed through the (frozen) source model to obtain internal representations. The target model is then trained to minimize the MSE between its features and those of the source model separately for source and target domain inputs. No labels are used, and the source model is not updated during this step. However, its BN layers adapt their running statistics, thereby internalizing the target domain distribution.
\end{enumerate}
While it is possible to implement the algorithm's steps in alternative ways, such as updating BN statistics via additional forward passes during the supervised step, combining statistics adaptation and feature distillation in a single pass offers greater computational efficiency. These two training steps implement our dual feature refinement strategy: Target domain statistics adaptation (see Section 2.2) and feature distillation (see Section 2.3).
Together, they align the feature spaces of both domains without requiring target domain labels.
During inference, we attach the source model's classifier head $g_s$ to the target model's feature extractor. Because the feature spaces are now aligned, the source model’s classifier can produce valid logits for target domain inputs. This constitutes the hypothesis transfer from the source to the target domain (see Section 2.4).
As long as the two domains exhibit semantic similarity—despite differences in style, presentation, or depiction—$DAFR^2$’s feature alignment enables effective and label-free hypothesis transfer across domains. 
Algorithm 1 outlines the steps of the $DAFR^2$ training procedure. The primary advantages of this algorithm lie in its simplicity, straightforward implementation, and the absence of additional hyperparameters beyond those typically required for training, such as the learning rate and weight decay, which are task-dependent. Furthermore, $DAFR^2$ is very stable and does not suffer from mode collapsing like many domain adaptation algorithms \cite{48.DA_Mode_Collapse1, 49.DA_Mode_Collapse2, 50.DA_Mode_Collapse2}.   

A key advantage of $DAFR^2$ is its ability to build a single model that performs well across both the source and target domains. This model not only achieves at least the same performance as a model trained with supervised learning on the source domain, but also demonstrates robustness to image corruptions. This dual-domain proficiency distinguishes $DAFR^2$ from other approaches that prioritize target domain performance at the expense of accuracy on the source domain. By excelling in both domains, the model intrinsically demonstrates its robust quality.

The three steps of the algorithm are essential in the domain adaptation process, and excluding any one of them leads to poor final results. While the step of adapting to the target domain's data distribution via BN statistics adaptation can yield performance improvements on its own, the other two steps are not impactful if used without it. The effects of feature distillation and hypothesis transfer heavily rely on the BN's statistics adaptation. Feature distillation, in particular, builds upon the adaptation of the source model to the principal components shared by the source and target domain data and amplifies the principal signals shared by both domains. Without the BN adaptation step, these signals are obscured by intricate correlations and are diffused across a vast, noisy feature space, rendering the subsequent steps ineffective.

\begin{algorithm}[h]
	\caption{The DAFR\textsuperscript{2} Algorithm}
	\begin{algorithmic}[1]
		\Function{TRAIN}{}
		\State Source and target models in training mode:
		\State \hspace{1em} $f_s.\text{train}(),\ g_s.\text{train}(),\ f_t.\text{train}()$
		\For{$\text{epoch} = 1$ to $MaxEpoch$}
		\State \textbf{Step 1:}
		\State Sample $x \in \mathcal{X},\ y \in \mathcal{Y}$
		\State Compute $g_s(f_s(x))$
		\State Compute $\mathcal{L}_{CE} = \text{CE}(g_s(f_s(x)), y)$
		\State Back-propagate grads for $\mathcal{L}_{CE}$, update $f_s$, $g_s$
		\State \textbf{Step 2:}
		\State Sample $x \in \mathcal{X},\ z \in \mathcal{Z}$
		\State Compute $f_s(x),\ f_s(z)$ \Comment{BN stats update}
		\State Compute $f_t(x),\ f_t(z)$
		
		\State Compute $\mathcal{L}_{\text{Regression}} = {}$$\frac{1}{2} \text{MSE}(f_t(x), f_s(x)) + {}$
		\Statex \hspace{\algorithmicindent} \hspace{\algorithmicindent} \hspace{\algorithmicindent} \hspace{\algorithmicindent} \hspace{\algorithmicindent} \hspace{\algorithmicindent} \hspace{\algorithmicindent}
		$\frac{1}{2} \text{MSE}(f_t(z), f_s(z))$
		\State Back-propagate grads for $\mathcal{L}_{\text{Regression}}$, update $f_t$
		\EndFor
		\EndFunction
		
		\Function{INFERENCE}{$x$}
		\State Put $f_t$ and $g_s$ in evaluation mode such as:
		\State $f_t.\text{eval}(),\ g_s.\text{eval}()$
		\State Predict $\hat{y} = g_s(f_t(x))$
		\EndFunction
	\end{algorithmic}
\end{algorithm}

\section{Experiments}
We evaluated $DAFR^2$ on a suite of popular benchmark datasets for robustness, MNIST-C \cite{51.MNIST-C}, CIFAR10-C \cite{4.BenchmarkingNNRobustness2CorruptionsPerturbations}, and CIFAR100-C \cite{4.BenchmarkingNNRobustness2CorruptionsPerturbations} and on a real-world medical image dataset,  PatchCamelyon-C \cite{52.PatchCamelyon-C}. The latter provides a critical real-world test case, as natural corruptions from the image acquisition process severely degrade the accuracy of models trained on clean data, causing them to perform only marginally better than random on this dataset. The simplicity of MNIST-C was also particularly useful, as it facilitates the visualization of 2D feature representations for in-depth algorithmic analysis, as discussed in Section 5.\par
All datasets are corrupted versions of the original respective datasets, created to evaluate the robustness of machine learning models (especially deep neural networks) to common visual corruptions and perturbations. The $"-C"$ suffix in these dataset names indicates that common corruptions (e.g., noise, blur, weather effects, digital distortions, etc.) have been artificially applied to the images of these datasets at various severity levels (5 severity levels in total – severity 0 being a random mild corruption and level 5 being a heavy random corruption). All corruptions and their descriptions are listed in Appendix A. These corrupted datasets are crucial for benchmarking and understanding how well machine learning models generalize to real-world, noisy, or slightly altered data, which is a key aspect of model robustness and out-of-distribution generalization.

\subsection{Experimental Setup}
Our experiments employ simple model architectures since our task is to study the effectiveness of $DAFR^2$ on the domain adaptation task rather than achieving state-of-the-art results on the original data. Specifically, we utilize a Residual Network 18 (ResNet18) \cite{54.ResNets} for all experiments. For feature distillation, we extract output from a linear layer appended after the average pooling layer of the ResNet model, just before the final linear classifier. This added linear layer matches the size of the pooling layer's output. Aside from this addition, the models remain otherwise unmodified. We train the source models using the Stochastic Gradient Descent (SGD) optimizer, while the target models are optimized with AdamW \cite{56.AdamW}. AdamW is a variant of the Adam \cite{55.Adam} optimizer that decouples weight decay from the gradient-based update, leading to better generalization and more principled optimization. Both models employ a cosine annealing scheduler \cite{57.CosineAnnealing} to adjust the learning rate during training. A comprehensive description of our experimental setups can be found in Appendix B.

\subsection{Results}
\begin{figure}[h!] 
	\includegraphics[width=0.47\textwidth]{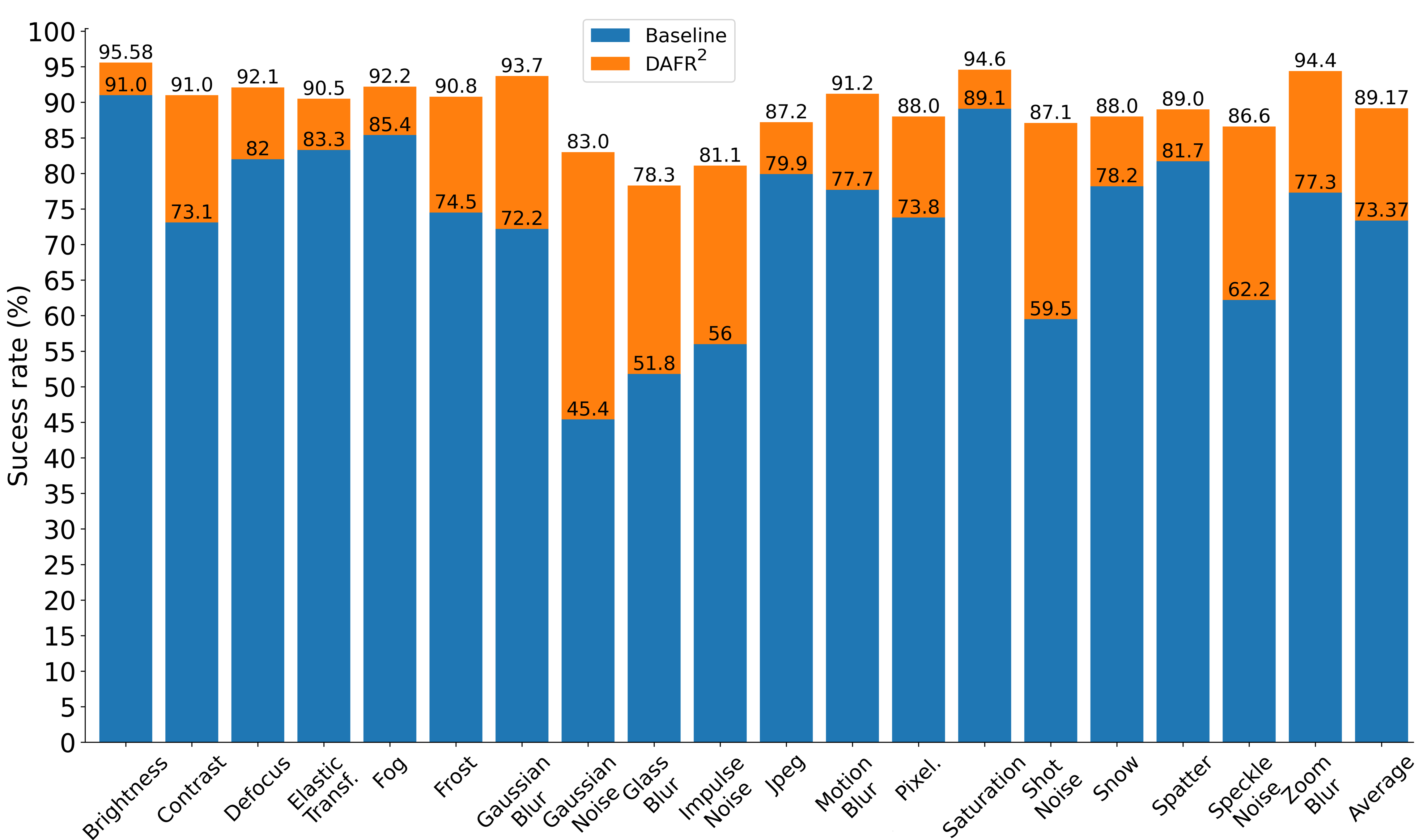} 
	\caption{$DAFR^2$'s results on CIFAR10-C per corruption type. Our algorithm shows improved results on CIFAR10-C, with a particularly substantial increase in success rate for corruption types where the model trained on clean data performs poorly, e.g., Gaussian noise, impulse noise and glass blurring.}
	\label{fig:CIFAR10-C results} 
\end{figure}

\textbf{Cifar10-C:} DAFR² was evaluated across all 19 distinct corruption types (at 5 different severity levels) present within the dataset. As a baseline, we consider the classification success rate obtained when using the model trained on the original (clean) data, which comprises the source domain of the problem. Specifically, we train a ResNet18 model in a supervised manner on clean data (CIFAR10) and test its performance on the corrupted images (CIFAR10-C). Our baseline results are on par with the baselines reported by previous works. DAFR² achieves a significantly improved success rate on the corrupted images compared to the baseline, while slightly improving the performance on the source domain, increasing it from $94.4\%$ to $94.8\%$. Especially for certain corruptions such as the Gaussian noise, the improvement is more than 35\%. Figure \ref{fig:CIFAR10-C results} shows the classification success rate and the improvement achieved by $DAFR^2$ on the Cifar10-C dataset per data corruption type, averaged over 5 runs for each corruption.

Although $DAFR^2$ is a training-time adaptation algorithm, it demonstrates superior performance compared to several test-time adaptation algorithms, such as the widely-used Test-time Entropy Minimization (TENT) algorithm \cite{58.TENT}. Crucially, this performance gain is achieved without compromising the model's accuracy on the original, clean data. Table \ref{tab:Cifar10-C comparison} provides a detailed comparison of $DAFR^2$ against other algorithms representing a diverse range of domain adaptation strategies. We must note that most of these works use bigger models than the ResNet18 we use, e.g., the WideResNet-28 \cite{59.WideResNets} and ResNeXt-29 \cite{60.ResNeXt} models.

\begin{table}[ht]
	\centering
	\caption{Comparative results between $DAFR^2$ and other state-of-the-art methods for CIFAR10-C.}
	\begin{tabular}{lccc}
		\hline
		\textbf{Method} & \textbf{Average Error(\%) $\downarrow$} \\
		\hline
		TTA \cite{102.TTA}    & 19.9  \\
		MEMO \cite{89.MEMO}  & 19.6  \\
		TTT-episodic \cite{95.TTT}  & 21.5 \\
		TTT \cite{95.TTT}   & 15.6 \\
		Tent \cite{58.TENT}  & 14   \\
		ETA \cite{91.ETA_EATA}   & 19.4 \\
		EATA \cite{91.ETA_EATA}  & 19.7 \\
		RoTTA \cite{103.RoTTA} & 25.2 \\
		RMT \cite{104.RobustMeanTeacher}   & 12.5 \\
		TEA \cite{105.TEA}   & 16.67 \\
		SAR \cite{28.Batch-InstanceNormalization}   & 20.23 \\
		\hline
		$DAFR^2$ & \textbf{10.83} \\
		\hline
	\end{tabular}
	\label{tab:Cifar10-C comparison}
\end{table}
More detailed experimental results are available in Appendix C.\\

\begin{figure}[h!] 
	\includegraphics[width=0.47\textwidth]{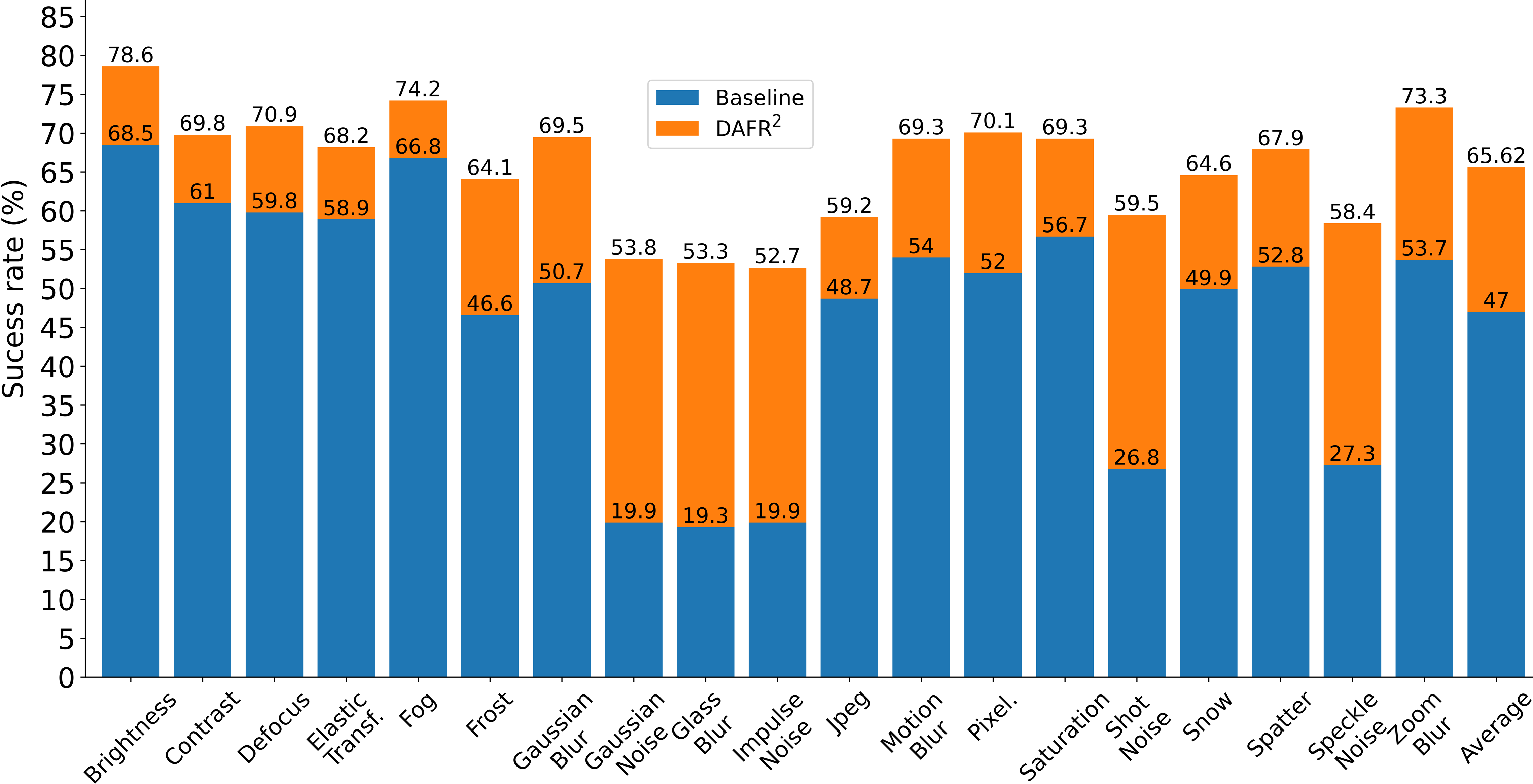} 
	\caption{The performance of $DAFR^2$ on CIFAR100-C, evaluated per corruption type, mirrors the findings on CIFAR10-C. Notably, our algorithm yields a significant improvement in the success rate for corruption types that cause a notable performance degradation in models trained on clean data.}
	\label{fig:CIFAR100-C results} 
\end{figure}

\noindent\textbf{Cifar100-C:} We evaluated our algorithm on CIFAR100-C, comparing its performance to a baseline model trained on CIFAR100. Consistent with our previous findings, $DAFR^2$ significantly improves performance on corrupted data while maintaining the model's accuracy on clean data (the baseline and adapted models achieved success rates of $72.3\%$ and $72.2\%$, respectively on the source domain testing set). Figure \ref{fig:CIFAR100-C results} shows the classification success rate obtained by our algorithm and the improvement achieved compared to the baseline. Table \ref{tab:Cifar100-C comparison} compares the results of $DAFR^2$ with state-of-the-art methods. Our method achieves superior adaptation results using a more compact architecture, the ResNet-18 model, while all other methods employ larger models like WideResNet28-10 and WideResnNet40-2A \cite{59.WideResNets}. Furthermore, unlike our approach, some of the methods used for comparison are based on online continual test-time adaptation.

\begin{table}[ht]
	\centering
	\caption{Comparative results between $DAFR^2$ and other state-of-the-art methods on CIFAR100-C. The suffix '-cont' denotes the continual test-time adaptation setting.}
	\begin{tabular}{lccc}
		\hline
		\textbf{Method} & \textbf{Average Error(\%)} $\downarrow$ \\
		\hline
		Shot \cite{44.NeuronSelectivityTransfer}   & 43.47  \\
		Tent \cite{58.TENT}   & 36.91  \\
		Tent-cont \cite{58.TENT} & 37.5 \\
		ETA \cite{91.ETA_EATA}    & 40.18 \\
		EATA \cite{91.ETA_EATA}   & 39.76 \\
		SAR \cite{37.TowardsStableTTAinDynamicWorld}    & 37.05   \\
		TEA \cite{105.TEA}    & 34.5 \\
		SAR-cont \cite{37.TowardsStableTTAinDynamicWorld}   & 35 \\
		CoTTa-cont \cite{106.CoTTA} & 38.2 \\
		\hline
		$DAFR^2$ & \textbf{34.38} \\
		\hline
	\end{tabular}
	\label{tab:Cifar100-C comparison}
\end{table}
More detailed experimental results are available in Appendix C.\\ 

\begin{figure}[h!] 
	\includegraphics[width=0.47\textwidth]{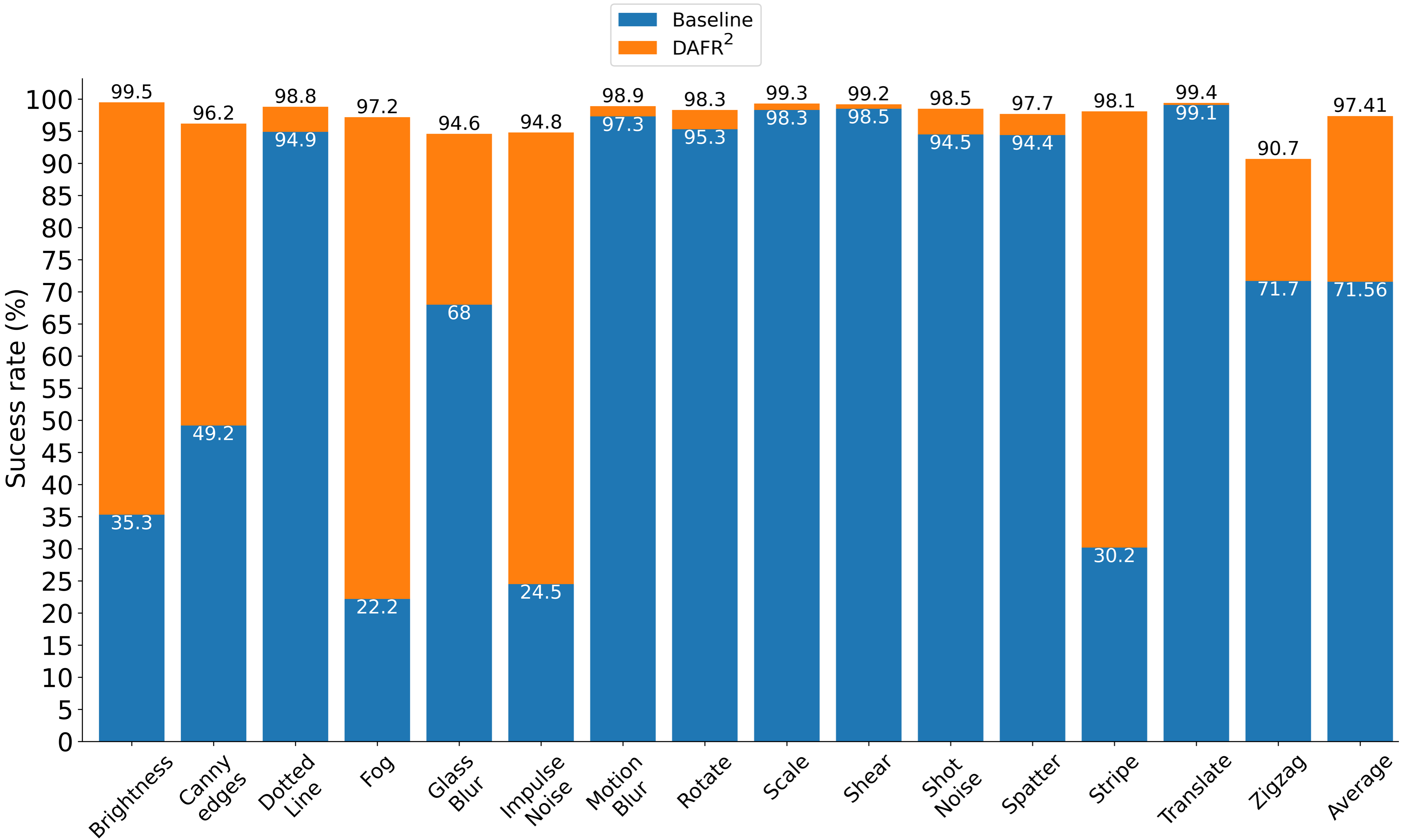} 
	\caption{$DAFR^2$'s results on MNIST-C per corruption type.}
	\label{fig:MNIST-C results} 
\end{figure}

\noindent\textbf{MNIST-C:} We applied $DAFR^2$ to the MNIST-C dataset and present the results in Figure \ref{fig:MNIST-C results}. Our baseline results were slightly lower than those reported by \cite{61.SourceFreeAdaptationToMeasurementShift}. Specifically, they achieved an average classification success rate of $72.3\%$, whereas our baseline yielded $71.54\%$. This performance drop primarily stems from our baseline model's poor performance on brightness and fog corruptions when trained solely on source domain data. A possible explanation for this discrepancy is our use of a ResNet18 model, which is significantly larger than the LeNet-5 architecture employed in the referenced work and may be more susceptible to overfitting specific data categories. Nevertheless, our baseline results on certain corruptions, like motion blur and translation, are significantly higher. 

A comparison of the performance of our algorithm with other works is shown in Table \ref{tab:MNIST-C comparison} and more detailed experimental results are available in Appendix C.

\begin{table}[ht]
	\centering
	\caption{Comparative results between $DAFR^2$ and other state-of-the-art methods on MNIST-C.}
	\begin{tabular}{lccc}
		\hline
		\textbf{Method} & \textbf{Average Error(\%)} $\downarrow$ \\
		\hline
		AdaBN \cite{82.AdaBN}  & 5.8  \\
		Shot-IM \cite{107.SHOT-IM} & 2.8  \\
		FR \cite{61.SourceFreeAdaptationToMeasurementShift}     & 3.3 \\
		BUFR \cite{61.SourceFreeAdaptationToMeasurementShift}    & 3.6 \\
		\hline
		$DAFR^2$ & \textbf{2.59} \\
		\hline
	\end{tabular}
	\label{tab:MNIST-C comparison}
\end{table}

\vspace{10pt}
\noindent\textbf{PatchCamelyon-C:} Besides famous benchmark datasets, we applied $DAFR^2$ on a dataset used for real-world medical imaging applications. The dataset focuses on the task of detecting metastatic tissue in histopathological scans. Specifically, the PatachCamelyon dataset addresses the important clinical problem of identifying metastatic breast cancer in lymph node sections, which is crucial for cancer staging and treatment and contains patches extracted from histopathologic whole-slide images (WSIs) of sentinel lymph node sections. WSIs were acquired at $40\times$ objective magnification ($0.243$ microns/pixel) and then undersampled at $10\times$ to increase the field of view. The dataset has a balanced distribution (approximately $50/50$) of positive and negative examples across all splits. Each patch in the dataset is $96\times96$ pixels with $3$ color channels (RGB) and is assigned a positive or negative label depending on whether at least one pixel of metastatic tumor tissue exists within the central $32\times32$ pixel region of the patch.
\par
\begin{figure}[h!] 
	\includegraphics[width=0.47\textwidth]{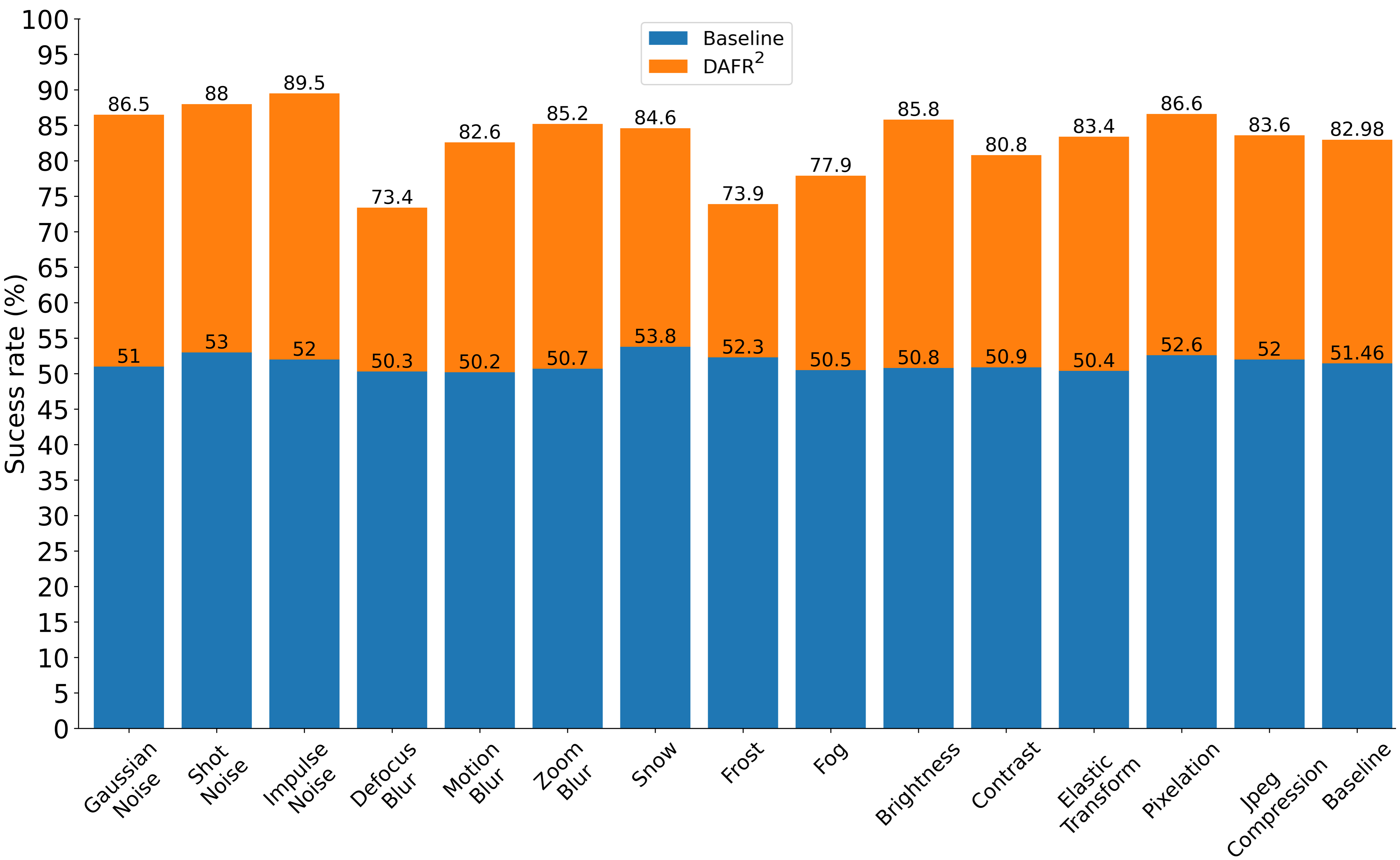} 
	\caption{$DAFR^2$'s results on the PatchCamelyon-C medical imaging dataset per corruption type.}
	\label{fig:PatchCamelyon-C results} 
\end{figure}

Figure \ref{fig:PatchCamelyon-C results} illustrates that the baseline model's performance is only marginally better than random chance, whereas $DAFR^2$ demonstrates a substantial improvement.

\section{Analysis}
This section analyzes our approach to understand how it achieves state-of-the-art performance despite its reliance on simple underlying concepts. Specifically, we examine the algorithm’s effect on the adapted feature space and quantify the enrichment of feature information during adaptation. To this end, we conducted $5$ experiments designed to elucidate $DAFR^2$'s contribution to enhancing domain adaptation. These experiments aim to:
\begin{enumerate}
	\item \textbf{Compare feature space geometry} before and after domain adaptation. We anticipate that $DAFR^2$ will align the source and target domain feature spaces.
	\item \textbf{Quantify mutual information (MI)} between source and target domain data, pre- and post-adaptation. We expect the algorithm to increase shared information substantially.
	\item \textbf{Compute the Fréchet Distance} \cite{62.FrechetDistance} between source and target domain data representations, before and after adaptation. We hypothesize $DAFR^2$ will decrease this distance, implying greater semantic similarity.
	\item \textbf{Calculate the local Lipschitz constant (LLC)} \cite{63.LocalLipschitzConstant} for both the baseline model (classifier trained on source data via supervised learning, without adaptation) and the $DAFR^2$-adapted model. We expect $DAFR^2$ to yield a lower local Lipschitz constant, which typically indicates improved robustness to distribution shifts and better generalization.
	\item \textbf{Investigate the impact on Cross-Entropy (CE) loss and feature-space reconfiguration}. $DAFR^2$ is expected to significantly reduce CE loss for corrupted data points that are misclassified by the baseline model. We also anticipate it will decrease the feature-space distance between these incorrectly classified points and other data points within their respective class. 
\end{enumerate}

\subsection{Feature Space Alignment}
\begin{figure}[h!] 
	\includegraphics[width=0.47\textwidth]{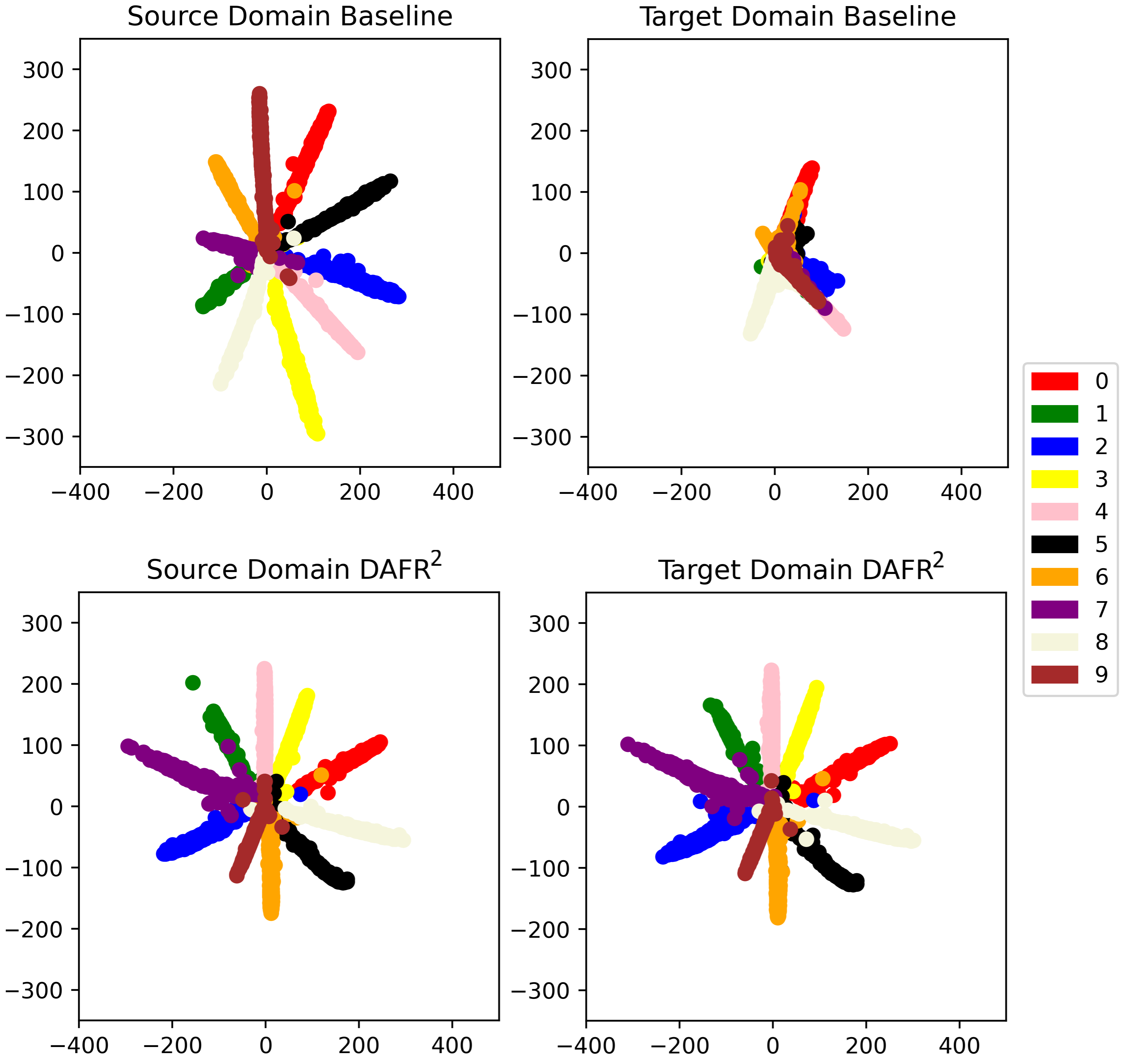} 
	\caption{Comparison of feature spaces for the source (left column) and target (right column) domains. The top row displays the feature spaces of the baseline model, which are misaligned without adaptation. In contrast, the bottom row demonstrates that the model trained with $DAFR^2$ achieves strong feature space alignment. The colormap links each digit class with a color. Best viewed in color.}
	\label{fig:feature_space} 
\end{figure}

To understand the impact of $DAFR^2$, this analysis compares the feature spaces of two models. First, we analyze a baseline model trained exclusively on source domain data. Second, we examine a model trained using $DAFR^2$. For both models, we input source and target domain data and observe the resulting feature spaces. We use features from models trained on the MNIST and MNIST-C datasets due to their simplicity, which allows us to reduce the feature dimensionality to 2 dimensions for visualization purposes without significantly compromising performance. The features obtained are shown in Figure \ref{fig:feature_space}. Without domain adaptation, the markedly dissimilar feature spaces directly impair the baseline model's performance when classifying target domain data. In contrast, when $DAFR^2$ performs domain adaptation by transferring knowledge from the baseline model to the adapted model, it aligns the two feature spaces, making them indistinguishable from each other. Feature space alignment enables the model trained with $DAFR^2$ to achieve significantly better performance on target domain data.

\subsection{Mutual Information-Based Representation Enhancement}
MI is a fundamental concept from information theory that measures the amount of information one random variable contains about another. When applied to two vectors (which represent two random variables or sets of observations), mutual information quantifies how much knowing one vector reduces uncertainty about the other. Mutual information can be used to encourage agreement between augmented views or stability of predictions, making models more robust to distribution shifts.
\par
In this analysis, we quantify the MI between feature representations of the source and target domain data. To evaluate the robustness of $DAFR^2$ to data corruptions, we compare the MI between the two domains using a model trained solely on the source domain and a model trained with $DAFR^2$. Estimating the MI of two random variables $x,y \in R^d$ is difficult because it requires joint and marginal distributions, i.e., knowing both $P(x,y)$ and $P(x)$, $P(y)$. For large values of $d$, traditional MI estimation methods (e.g., histograms and kernel density estimation) suffer from the curse of dimensionality and become highly unreliable \cite{64.MINE}. Another reason computing mutual information is difficult is that it relies on all types of dependencies—both linear and non-linear—making correlation-based estimators inherently inadequate \cite{65.On_MI_estimation}. To overcome these problems in estimating the MI between the representations we use Mutual Information Neural Estimation (MINE) \cite{64.MINE}. MINE learns a neural network that distinguishes between joint and marginal samples, effectively learning to maximize MI by pushing the network to assign higher scores to joint pairs by approximating the Donsker–Varadhan (DV) \cite{66.Donsker-Varadhan} lower bound on the Kullback–Leibler divergence between the joint distribution and the product of the marginals. It's especially useful for high-dimensional or continuous variables where traditional MI estimation fails. We train the MINE network using representations from the MNIST and MNIST-C datasets, obtained with the original setup: a ResNet18 backbone for both domains and a representation size of 128. Table \ref{tab:MI} shows the MI estimations for several corruptions for both the scenarios of using $DAFR^2$ and using the model trained on source domain data only. 
\begin{table}[ht]
	\centering
	\caption{MI comparisons between the representations of source and target domains with and without applying $DAFR^2$.}
	\begin{tabular}{lccc}
		\hline
		\textbf{Distortion} & \textbf{Source Only} & \bm{$DAFR^2$} & \shortstack{\\ \textbf{\%$\uparrow$}} \\
		\hline
		Brightness    & 2.24 & 6.12 $\uparrow$ & 273 \\
		Canny edges   & 2.67 & 3.49 $\uparrow$ & 131 \\
		Dotted lines  & 3.62 & 4.41 $\uparrow$ & 122 \\
		Fog           & 1.34 & 3.71 $\uparrow$ & 277 \\
		Glass Blur    & 2.33 & 3.26 $\uparrow$ & 140 \\
		Impulse Noise & 1.31 & 3.15 $\uparrow$ & 240 \\
		Spatter       & 2.98 & 3.59 $\uparrow$ & 121 \\
		Stripe        & 2.40 & 3.51 $\uparrow$ & 146 \\
		Zigzag        & 2.76 & 3.26 $\uparrow$ & 118 \\
		\hline
	\end{tabular}
	\label{tab:MI}
\end{table}
We observe that MI increases consistently and substantially with the application of $DAFR^2$. This suggests that the representations computed from the data of the two domains share structure and are more effectively aligned. Elevated MI implies that the learned representations preserve more domain-invariant characteristics. This facilitates improved generalization across domains, as the model leverages shared underlying structure. The increase in MI suggests a more robust and informative feature encoding which contributes to the observed performance gains under domain shift scenarios.

\subsection{Fréchet Distance}
The Fréchet Distance (FD) between representations, especially in the context of neural network embeddings, is commonly used to quantify the similarity between two multivariate Gaussian distributions fit to sets of feature vectors (e.g., from different domains, models, or layers). This is often referred to as the Fréchet Inception Distance (FID) in computer vision. To evaluate the effect of our algorithm on the quality of learned representations, we compute the FID between feature distributions computed on the model trained with source domain data only and the model trained with $DAFR^2$. Table \ref{tab:fid} shows the results of this analysis using the CIFAR10 and CIFAR10-C datasets. We selected these datasets to use a more complex alternative to MNIST for this experiment.
\begin{table}[ht]
	\centering
	\caption{FID between the representations of source and target domains with and without applying $DAFR^2$.}
	\begin{tabular}{lccc}
		\hline
		\textbf{Distortion} & \textbf{Source Only} & \bm{$DAFR^2$} & 
		\shortstack{\\ \textbf{Reduction} \\ \textbf{Factor ($\times$)}} \\
		\hline
		Brightness     & 6243.00    & 8.90 $\downarrow$     & 701.40 \\
		Canny edges    & 1991.00    & 39.80 $\downarrow$    & 50.00 \\
		Dotted lines   & 86.13      & 4.77 $\downarrow$     & 18.00 \\
		Fog            & 13664.00   & 43.32 $\downarrow$    & 315.40 \\
		Glass Blur     & 1841.42    & 17.15 $\downarrow$    & 107.37 \\
		Impulse Noise  & 6209.63    & 49.11 $\downarrow$    & 126.44 \\
		Spatter        & 274.16     & 28.77 $\downarrow$    & 9.53 \\
		Stripe         & 3881.76    & 119.94 $\downarrow$   & 32.36 \\
		Zigzag         & 340.42     & 41.43 $\downarrow$    & 8.21 \\
		\hline
	\end{tabular}
	\label{tab:fid}
\end{table}
The FID decreases significantly when applying $DAFR^2$, supporting our hypothesis that the feature space becomes better aligned, the representations more effectively capture the underlying structure, and are less influenced by subtle domain-specific characteristics.

\subsection{Local Lipschitz Constant}
A low LLC for a model (typically a neural network) indicates that the model's output is not highly sensitive to small changes in the input, at least in a neighborhood around a specific input point. We hypothesize that $DAFR^2$ trains a target domain model with a lower LLC, resulting in enhanced generalization, smooth gradient flowing and smooth local behavior under distribution shifts. 

We estimate the LLC of models trained on CIFAR10 and CIFAR10-C by evaluating their sensitivity to random perturbations. Specifically, we sample 200,000 random inputs from a standard Gaussian distribution, each matching the input dimensions of the dataset ($32 \times 32 \times 3$). For each sample, we compute the gradient of the $L_2$ norm of the model’s output with respect to the input. The maximum gradient norm observed across all samples is then used as the LLC estimate. Table \ref{tab:lipschitz} shows the LLC of the model trained with the data of the source domain (baseline) and tested on the target domain (all corruptions) and the LLC estimates per corruption for models trained with $DAFR^2$.

We observe that the LLC estimates of models trained with $DAFR^2$ are substantially lower than those of the baseline model. This suggests that our method effectively smooths the model’s local behavior, making it less sensitive to distribution shifts that preserve the semantic content of the images.

\begin{table}[h]
	\centering
	\caption{Lipschitz constant measured for a model trained only on the source domain data and a model trained with $DAFR^2$.}
	\begin{tabular}{lccc}
		\hline
		\textbf{Model} & \textbf{LLC} & \shortstack{\\ \textbf{Reduction} \\ \textbf{Factor ($\times$)}} \\
		\hline
		\textbf{Source Domain only} & 6.04 & -- \\
		\hline
		\bm{$DAFR^2$} & &\\
		Brightness           & 0.76 $\downarrow$ & 7.95 \\
		Elastic Transform    & 0.69 $\downarrow$ & 8.75 \\
		Glass Blur           & 0.66 $\downarrow$ & 9.15 \\
		Contrast             & 1.76 $\downarrow$ & 3.43 \\
		Defocus Blur         & 1.44 $\downarrow$ & 4.19 \\
		Gaussian Noise       & 3.80 $\downarrow$ & 1.59 \\
		Impulse Noise        & 2.32 $\downarrow$ & 2.60 \\
		Speckle              & 2.76 $\downarrow$ & 2.19 \\
		Shot Noise           & 2.93 $\downarrow$ & 2.06 \\
		Zoom Blur            & 2.23 $\downarrow$ & 2.70 \\
		Gaussian Blur        & 1.16 $\downarrow$ & 5.21 \\
		Average              & 1.86 $\downarrow$ & 3.25 \\
		\hline
	\end{tabular}
	\label{tab:lipschitz}
\end{table}

\subsection{Impact on Cross-Entropy (CE) loss and feature-space reconfiguration}
\par
We also conducted two additional experiments investigating how our algorithm impacts 
\begin{enumerate}
	\item The CE loss of corrupted data points compared to the CE loss experienced with the baseline model, i.e., the model trained solely on data from the source domain. 
	\item The distance between the representations of corrupted data points that are incorrectly classified by the baseline and the nearest same-class representation.
\end{enumerate}
Appendix D provides a full description of these experiments and their outcomes. Concretely, $DAFR^2$ significantly decreases both the observed CE losses and the observed distances between data point representations that are relevant to the experiment.

\section{Related Work}
In this section, we review related work in domain adaptation and discuss how our approach aligns with or diverges from existing methods. We particularly discuss domain adaptation in general, feature refinement and normalization-based adaptation. 

\subsection{Domain Adaptation}
Domain adaptation has been a pivotal research area since the very beginning of machine learning and neural networks. There is a fundamental connection between domain adaptation and generalization because closing the "domain gap"—the difference between training data and real-world data—significantly boosts a model's ability to perform well in diverse environments. This crucial link has spurred extensive research into various domain adaptation techniques, resulting in a wealth of diverse methodologies. 

The major domain adaptation families differ in the supervision type, the adaptation strategy and the type of alignment they apply. Based on supervision type, the methods can be divided into unsupervised \cite{67.Domain-AdversarialTrainingNN, 68.MCD_UnsupervisedDA, 47.DoWeReallyNeed2AccessSourceData}, semi-supervised \cite{70.TriTrainingUnsupDA, 71.FixBi} and supervised \cite{72.DeepCoral, 73.TransferableCurriculumLearning} domain adaptation. Based on the strategy applied, we have  discrepancy-based methods \cite{72.DeepCoral, 74.JointAdaptationNets, 75.BridgingTheoryAlgorithmsDA}, adversarial-based methods \cite{67.Domain-AdversarialTrainingNN,76.CDAN, 77.DADDA}, reconstruction-based methods \cite{78.DomainSeparationNetowrks, 79.UnsupervisedPixel-LevelDA}, contrastive methods or self-supervised methods \cite{80.ContrastiveAdaptationNetwork, 81.CrossdomainContrastiveLearning} and normalization-based methods \cite{82.AdaBN, 28.Batch-InstanceNormalization, 35.TheNormMustGoOn, 36.ImprovingRobustnessByCovariateShiftAdapt, 37.TowardsStableTTAinDynamicWorld}. Based on the type of alignment, we have feature-level alignment \cite{67.Domain-AdversarialTrainingNN, 83.AFN-AdaptiveFeatureNorm}, pixel-level adaptation \cite{84.CycleGAN, 85.SimGAN} and output-space adaptation \cite{86.ADVENT,87.CLAN,88.HRDA}. $DAFR^2$ lies within the unsupervised, feature-level domain adaptation regime and, due to its adaptation to target domain statistics, shares common ground with normalization-based methods.

In recent years, several emerging domain adaptation settings have achieved promising performance on the task. Test-time-adaptation (TTA) \cite{58.TENT, 89.MEMO} is an interesting approach to domain adaptation that focuses on adapting the model during inference using only target data, assuming no access to source data at test time. For example, TENT \cite{58.TENT}
adapts to unlabeled test data by minimizing prediction entropy, updating only the affine parameters of the BN layers, i.e., their scale and shift parameters. By freezing all other model parameters, TENT reduces the risk of catastrophic forgetting and improves efficiency by limiting the number of required updates. Another interesting TTA method is Source Hypothesis Transfer (SHOT) \cite{47.DoWeReallyNeed2AccessSourceData}, which aligns features from the source domain to the target domain using hierarchical clustering and optimal transport theory to minimize the distributional discrepancy. In the same context, SAR \cite{37.TowardsStableTTAinDynamicWorld} minimizes the entropy of the model's predictions and employs sharpness-aware minimization (SAM) \cite{90.SAM} to encourage the model to converge to flatter minima in the loss landscape, making it more robust to small perturbations at test time. Efficient Test-Time Adaptation (EATA) \cite{91.ETA_EATA} employs selective entropy minimization to encourage adaptation only on reliable and non-redundant test samples, which are identified through an active sample selection criterion. Furthermore, EATA also implements a Fisher Regularization to prevent forgetting previously learned knowledge. For a comprehensive overview of test-time domain adaptation methods, we refer the reader to a survey by Liang et al. on TTA under distribution shifts \cite{92.SurveyDA}.  

Similar to test-time adaptation, Test-Time Training approaches \cite{93.MT3,94.TTTFlow} aim to improve model robustness during inference without relying on labeled target data. TTT \cite{95.TTT}, for example, introduces a self-supervised objective at test time: It tasks the model with predicting the rotation applied to each input image. This encourages the model to learn rotation-invariant features, enhancing generalization under distribution shifts. TTTFlow \cite{94.TTTFlow}, on the other hand,   instead of rotation prediction, integrates optical flow prediction as the self-supervised task. The optical flow task encourages the model to learn motion-related spatio-temporal features that are more transferable across domains.

\par
Some simple but highly effective strategies rely on more conventional methods to achieve competitive performance, often employing well-established, data-driven techniques such as data augmentation. Augmix \cite{22.Augmix} applies randomly chosen image augmentation sequences to produce a range of semantically sound image variations. For each original image, several chains with different depth and severity are generated using convex combinations sampled from a Dirichlet distribution. This mixing enhances diversity while maintaining the original image. Additionally, AugMix enhances regularization and model robustness by blending the original image with its augmented versions through a skip connection, using weights sampled from a Beta distribution. Rusak et al. \cite{22b.SimpleWay2MakeNNsRobust} introduce another simple way to make NNs robust against image corruptions and suggest the use of carefully calibrated additive Gaussian and speckle noise. Notably, their study demonstrates that simple noise-based augmentation, especially when combined with adversarial training against specific noise distributions, can substantially enhance the robustness of NNs.

\subsection{Feature Refinement}
Feature refinement improves the generalizability of models, particularly in challenging situations such as distribution shifts. Feature alignment focuses on transferring intermediate representations from one model to another and is derived from the feature distillation technique, originally proposed for model compression. In feature distillation, usually a smaller NN called the student is trained to mimic the unnormalized output (logits) of a larger teacher model. This task often leads to better performance and an effective transfer of knowledge from the teacher to the student model. Extending this concept to learning internal representations rather than the teacher's output, the resulting model can learn more meaningful and robust features. Gustavo Aguilar et al. \cite{96.KnowledgeDistillFromInternalRepresentations} highlight the limitations of traditional distillation methods that focus solely on output logits and suggest that aligning internal representations enhances student model generalization. They proved the effectiveness of their approach by refining the internal representations of a BERT model to a simplified one, achieving better generalization than using traditional knowledge distillation. 
As one of the first approaches to feature refinement in imagery, FitNets \cite{39:FitNets} introduced the concept of using an MSE loss for intermediate feature regression as 'hints' to steer the student model training. Yim et al. \cite{43.AGiftFromKnowledgeDistillation} extended this concept by applying an MSE regression loss to the Gram matrix formed by taking the inner product of the feature maps from two different layers of the source model. Furthermore, they demonstrated that models utilizing feature refinement exhibit cross-task knowledge transfer even when trained on various tasks.
\par
Beyond minimizing MSE, other methods utilize different losses to align the distribution of activations between two models. For example, Huang and Wang \cite{44.NeuronSelectivityTransfer} instead of matching features one-to-one, they minimize the Maximum Mean Discrepancy (MMD) between the feature distribution of the source model and the target model. This approach ensures that the target model exhibits activation statistics that match those of the source model. Tian et al. \cite{45.ContrastiveRepresentationDistillation} introduced a novel approach to feature alignment, employing a contrastive loss to maximize the mutual information between the feature spaces of two models. They propose applying this alignment at the final embedding layer (the penultimate layer before classification). This strategic placement ensures the alignment occurs on deep, highly representative features, leading to more effective knowledge transfer. $DAFR^2$ adapts this strategy and applies feature refinement on the penultimate layer. 
\par
All feature alignment methods described so far are executed in a supervised manner. More recently, self-supervised and unsupervised feature distillation methods have gained significant attention because of their remarkable performance and the ability to leverage vast amounts of unlabeled data. Boost your own latent (BYOL) \cite{97.BYOL} trains two instances of the same CNN (a teacher and a student model) with the teacher's weights being an exponential moving average of the student's weights and the feature alignment conducted with an MSE loss. The momentum teacher was trained in a self-supervised manner, and the regression of its representations with an MSE loss led to excellent results without the use of negative samples to prevent mode collapse during training. Finally, SimSiam \cite{98.SImSiam} employs two branches of the same network, both processing different augmented views of the same image. Unlike BYOL, SimSiam does not use a separate teacher model. Instead, it relies on a stop-gradient operation to prevent collapse. The model minimizes the negative cosine similarity between the predictions of the two branches, which is mathematically equivalent to minimizing the mean squared error (MSE) after feature normalization.  

\subsection{Normalization-based Adaptation}
The information captured by each feature map in a CNN can be divided into two key components: style (appearance statistics) and shape (spatial structure) \cite{28.Batch-InstanceNormalization, 99.StyleTransfer, 100.ImageStyleTransferCNNs}. The style of the feature maps is predominantly represented by the statistics (mean and variance) within the BN layers following the convolutional operations. These statistics encapsulate the texture and appearance of the features, while the spatial structure (shape) is preserved in the location and arrangement of activations across the feature map.
Nam et al. \cite{28.Batch-InstanceNormalization} propose the learning of a gating parameter per feature map channel that decides how much style information should be preserved or suppressed. Mirza et al. \cite{35.TheNormMustGoOn} suggest an online adaptation approach by creating a batch of augmentations for every target domain sample that arrives, which is used to adapt the running mean and variance of the BN layers using an adaptive momentum parameter. A much simpler approach by Nado et al. \cite{101.EvaluatingPredictionTimeBatchNormForRobustness} recomputes BN mean and variance on each test batch, resulting in both better accuracy and calibration metrics, i.e., Expected Calibration Error (ECE) and Brier score. Combining the training and test set BN statistics, Schneider et al. \cite{29.CovariateShiftAdaptation} in a context blend to balance stability and adaptation.
\par 
$DAFR^2$ effectively aligns the feature spaces of the source and target models by combining MSE–based feature alignment with adaptation to target domain statistics. The adaptation encourages the source model to structure its feature space around the dominant principal components shared across domains. Meanwhile, the MSE loss penalizes high-frequency oscillations—responsible for sharp local differences—more heavily than smooth variations around these components. As a result, the target model exhibits stable behavior under moderate distribution shifts, avoiding spurious responses.

\section{Discussion and Conclusions}
This work introduces the $DAFR^2$ framework which comprises a robust and effective strategy for domain adaptation by leveraging feature refinement through statistics alignment, feature distillation and hypothesis transfer. Unlike many existing approaches that rely heavily on complex objectives and require complex optimization schemes, $DAFR^2$ operates through conceptually simple yet powerful mechanisms. Its strong performance across multiple benchmark datasets underscores its effectiveness in bridging the distribution gap between source and target domain without relying on target labels or requiring complex model architectures.
\par
The three mechanisms employed by $DAFR^2$ act synergically to align internal feature representations. The adaptation of BN statistics enables the source model to internalize structural properties of the target domain. Feature distillation enables representation-level matching, suppresses domain-specific noise and shifts the focus on core, domain-invariant features. Hypothesis transfer allows the model to exploit the aligned representations at inference time by preserving decision boundaries learned on clean data and applying them to aligned target representations. This renders the use of target domain labels unesessary making $DAFR^2$ very practical for cases that such labels are unavailable. 
\par
Compared to test-time adaptation methods such as Tent or SHOT, $DAFR^2$ achieves superior or comparable performance with a training-time strategy that avoids the pitfalls of catastrophic forgetting and the instability of online updates. A significant advantage that $DAFR^2$ has over other domain adaptation methods, is that it does not sacrifice source domain performance.
\par
Despite its strengths, $DAFR^2$'s reliance on BN layers may limit its applicability to architectures that do not employ such layers. Additionally, while we show improvements across a range of corruption types, we do not investigate the method's effectiveness under extreme domain shifts or semantic drift which will be investigated in future work.
\par
$DAFR^2$ offers a compelling alternative for the domain adaptation task. Its simplicity and strong empirical results suggest it can be used as a foundational component for future work aiming to improve model robustness under domain shift.

{\small
	\bibliographystyle{plain}
	\bibliography{bibliography}}

\newpage
\onecolumn

\section*{APPENDICES}
\appendix
\section{Dataset corruptions}
\noindent In this appendix, we describe the various corruptions applied to the datasets used in our experiments. The corruptions are divided to four families (Noise, Blur, Weather and Digital corruptions).

\begin{table}[ht]
	\centering
	\caption{Corruptions used in the experimental datasets, their family type and a short description}
	\begin{tabular}{lll}
		\toprule
		\textbf{Corruption Type} & \textbf{Name} & \textbf{Description} \\
		\midrule
		\textbf{Noise} &
		Gaussian & Adds random values drawn from a Gaussian distribution to pixels \\
		& Shot & Simulates noise caused by light fluctuations (e.g., low light) \\
		& Impulse & "Salt-and-pepper" noise, pixels are randomly set to black or white \\
		& Speckle & High-frequency noise, generated by coherent imaging systems \\
		\midrule
		\textbf{Blur} &
		Defocus & Simulates out-of-focus camera lens effects \\
		& Glass & Simulates the effect of looking through glass \\
		& Gaussian & Applies a Gaussian filter to smooth the image \\
		& Motion & Simulates blur caused by camera or object movement \\
		& Zoom & Simulates blur that occurs when zooming during image capture \\
		\midrule
		\textbf{Weather} &
		Snow & Adds synthetic snow effects \\
		& Frost & Adds synthetic frost effects \\
		& Fog & Adds synthetic fog effects, reducing visibility \\
		& Brightness & Adjusts the overall brightness of the image \\
		\midrule
		\textbf{Digital} &
		Contrast & Adjusts the difference between light and dark areas \\
		& Elastic Transformation  & Applies local, non-linear elastic deformations \\
		& JPEG Compression & Simulates artifacts introduced by JPEG image compression \\
		& Pixelate & Reduces image resolution by grouping pixels into larger blocks \\
		& Spatter & Simulates splashes or smudges on the image \\
		& Saturate & Adjusts the color saturation of the image \\
		& Canny Edges & Applies the Canny edge detection algorithm, reducing to edge contours \\
		& Zigzag & Overlays zigzag patterns onto the image, distorting the original digit shapes \\
		& Dotted Line & Superimposes randomly oriented dotted lines, introducing occlusions \\
		& Rotate & Rotates the image by a certain degree causing orientation changes \\
		& Scale & Resizes the image up or down, simulating zoom effects \\
		& Sheer & Applies a shearing transformation, shifting the image in a specific direction \\
		& Stripe & Adds vertical stripes obscuring parts the image \\
		& Translate & Shifts the image position horizontally or vertically causing positional changes \\
		\bottomrule
	\end{tabular}
	\label{tab:Distortions}
\end{table}

\newpage 
\section{The experimental setup}
\noindent This appendix discusses in detail the experimental setup for applying $DAFR^2$ on various datasets. To test the $DAFR^2$ framework, we ran experiments on several standard datasets: MNIST-C, CIFAR10-C, CIFAR100-C, and PatchCamelyon-C. We used a ResNet18 architecture for both the source and target models in all our experiments. This allowed us to focus on the efficacy of the adaptation method, rather than on surpassing other state-of-the-art methods with more complex architectural features.
\par
The source model was trained using the SGD optimizer and a learning rate of $0.1$, while the target model was optimized with AdamW and a learning rate of $0.001$. AdamW was chosen as it is a variant of the Adam optimizer that decouples weight decay from the gradient-based update, which leads to better generalization and more principled optimization. Both models used a cosine annealing scheduler to adjust the learning rate during training, which helps with stable convergence. We set $T_{max}$ to $300$ and $eta_{min}$ to $0.0001$ and weight decay to $0.0001$ for all experiments.

\par
We employed the standard augmentations of random cropping with padding of $4$ pixels and random horizontal flipping with $p=0.5$ for all experiments. Finally, we applied normalization on the images based on the mean and variance computed on the training sets. 

Feature distillation, a key part of our approach, involves the addition of a linear layer right after the average pooling layer in the ResNet18 model. This new layer's output served as the feature embedding that the target model learned to replicate. The addition of a linear layer makes the model more flexible at computing expressive features shared by both domains, without relying on the rather cumbersome average pooling operation.

\vspace{30pt}
\section{Detailed Experimental Results}
\noindent In this appendix, we present more detailed results based on our experiments on CIFAR10-C, CIFAR100-C and MNIST-C. 

\begin{table}[ht]
	\caption{A comprehensive comparison between $DAFR^2$ and various state-of-the-art methods on CIFAR10-C and CIFAR100-C for all corruptions. While all other methods utilize the WRN-28-10 architecture with BN, $DAFR^2$ employs the more compact ResNet18 model. The best performance scores are highlighted in bold.}
	\centering
	\resizebox{\textwidth}{!}{%
		\begin{tabular}{@{}llcccccccccccccccc@{}}
			\toprule
			& \multirow{2}{*}{Method} & \multicolumn{3}{c}{Noise} & \multicolumn{4}{c}{Blur} & \multicolumn{4}{c}{Weather} & \multicolumn{4}{c}{Digital} & \multirow{2}{*}{Avg(↑)} \\ 
			\cmidrule(lr){3-5} \cmidrule(lr){6-9} \cmidrule(lr){10-13} \cmidrule(lr){14-17}
			& & Gaussian & Shot & Impulse & Defocus & Glass & Motion & Zoom & Snow & Frost & Fog & Bright & Contrast & Elastic & Pixel & JPEG & \\ \midrule
			\multirow{9}{*}{\rotatebox{90}{CIFAR10(-C)}} 
			
			& Source & 45.4 & 59.5 & 56 & 82 & 51.8 & 77.7 & 77.3 & 78.2 & 74.5 & 85.4 & 91 & 73.1 & 83.3 & 73.8 & 79.9 & 73.37 \\
			& TENT   & \textbf{89} & 86.2 & 89.2 & 86.7 & 89.2 & 74.3 & 87.9 & 77.4 & 88.2 & 80.1 & 87.4 & 87.3 & 89.0 & \textbf{88.3} & \textbf{89.7} & 86\\ 
			& RoTTA   & 73.2 & 66.4 & 60.5 & 76.4 & 60.4 & 81.9 & 84.9 & 78.3 & 78.1 & 81.2 & 88.8 & 83.5 & 69.3 & 66.3 & 72.2 & 74.76\\
			
			& ETA   & 72.21 & 73.88 & 63.72 & 87.21 & 64.70 & 85.84 & 87.89  & 82.73 & 82.58 & 84.77 & 91.6 & 87.32 & 76.22 & 80.37 & 72.69 & 79.58 \\ 
			& EATA  & 72.25 & 73.88 & 63.74 & 87.19 & 64.72 & 85.83 & 87.89 & 82.73 & 82.61 & 84.75 & 91.61 & 87.35 & 76.25 & 80.33 & 72.70 & 79.59\\ 
			
			& TEA & 78.33 & 79.87  & 70.94 & 88.89 & 71.31 & 87.87 & 89.77 & 85.56 & 85.29 & 87.61 & 92.37 & 88.98 & 79.32 & 84.9 & 78.99 & 83.33\\ 
			
			& SAR & 71.95 & 74.14 & 64.11 & 87.39 & 65.2 & 86 & 88.06 & 83.08 & 82.66 & 85.07 & 91.9 & 87.2 & 76.69 & 80.41 & 72.79 & 79.77\\ 
			\cmidrule(l){2-18} 
			& $DAFR^2$    & 83.0 & \textbf{87.1} & \textbf{81.1} & \textbf{92.1} & \textbf{78.3} & \textbf{91.2} & \textbf{94.4} & \textbf{88.0} & \textbf{90.8} & \textbf{92.2} & \textbf{95.58} & \textbf{91.0} & \textbf{90.5} & 88.0 & 87.2 & \textbf{89.17} \\ \midrule \midrule
			
			\multirow{10}{*}{\rotatebox{90}{CIFAR100(-C)}} 
			& Source & 19.9 & 26.8 & 19.9 & 59.8 & 19.3 & 54 & 53.7 & 49.9 & 46.6 & 66.8 & 68.5 & 61 & 58.9 & 52 & 48.7 & 47 \\
			& BN     & 46.53 & 48.62 & 37.15 & 70.94 & 47.36 & 69.04 & 71.25 & 63.00 & 62.96 & 66.08 & 75.89 & 71.31 & 58.79 & 64.56 & 47.46 & 60.06 \\
			& PL     & 29.54 & 34.09 & 14.99 & 64.06 & 40.81 & 65.36 & 67.74 & 60.42 & 59.47 & 62.92 & 75.98 & 57.29 & 58.72 & 59.41 & 50.34 & 53.40 \\
			& SHOT   & 37.50 & 39.68 & 21.27 & 67.90 & 45.52 & 67.42 & 69.86 & 61.42 & 60.75 & 65.01 & 75.12 & 63.96 & 58.96 & 62.54 & 51.09 & 56.53 \\
			& TENT   & 53.44 & 54.12 & 45.53 & 71.69 & 51.17 & 71.54 & 71.63 & 64.88 & 65.30 & 68.41 & 75.14 & 73.59 & 59.25 & 66.81 & 53.93 & 63.09 \\
			& ETA    & 48.64 & 50.95 & 38.05 & 69.66 & 47.52 & 67.76 & 70.24 & 62.51 & 61.73 & 66.03 & 73.40 & 71.15 & 56.93 & 64.40 & 48.34 & 59.82 \\
			& EATA   & 48.76 & 51.60 & 39.47 & 69.29 & 47.49 & 68.13 & 70.68 & 62.94 & 62.53 & 65.14 & 74.48 & 71.61 & 57.53 & 64.24 & 49.67 & 60.24 \\
			& SAR    & 51.34 & 54.08 & 44.62 & 72.24 & 50.10 & 71.06 & 72.43 & 64.96 & 65.35 & 68.40 & 76.23 & 73.95 & 60.04 & 67.26 & 52.29 & 62.95 \\ 
			& TEA    & \textbf{54.29} & 56.55 & 48.59 & \textbf{72.96} & \textbf{53.78} & \textbf{72.63} & \textbf{74.20} & \textbf{67.78} & \textbf{67.14} & 69.98 & 76.74 & \textbf{75.71} & 62.18 & 68.65 & 55.32 & 65.10
			\\ \cmidrule(l){2-18} 
			& $DAFR^2$   & 53.8 & \textbf{59.5} & \textbf{52.7} & 70.9 & 53.3 & 69.3 & 73.3 & 64.6 & 64.1 & \textbf{74.2} & \textbf{78.6} & 69.8 & \textbf{68.2} & \textbf{70.1} & \textbf{59.2} & \textbf{65.62} \\ \bottomrule
		\end{tabular}%
	}
	\label{tab:tea_comparison_large_margins}
\end{table}

\begin{table}[ht]
	\caption{A comprehensive comparison between $DAFR^2$ and various state-of-the-art methods on the MNIST-C dataset for all corruptions. The best performance scores are highlighted in bold.}
	\centering
	\resizebox{\textwidth}{!}{%
		\begin{tabular}{@{}llcccccccccccccccc@{}}
			\toprule
			& \multirow{2}{*}{Method} & \multicolumn{2}{c}{Noise} & \multicolumn{9}{c}{Digital} & \multicolumn{2}{c}{Weather} & \multicolumn{2}{c}{Blur} & \multirow{2}{*}{Avg(↑)} \\ 
			\cmidrule(lr){3-4} \cmidrule(lr){5-13} \cmidrule(lr){14-15} \cmidrule(lr){16-17}
			& & Impulse & Shot & Spatter & Canny & Dotted & Shear & Scale & Rotate & Stripe & Translate & ZigZag & Fog & Brightness & Glass & Motion & \\ \midrule
			
			\multirow{6}{*}{\rotatebox{90}{MNIST(-C)}} 
			
			& Source & 24.5 & 94.5 & 94.4 & 49.2 & 94.9 & 98.5 & 98.3 & 95.3 & 30.2 & 99.1 & 71.7& 22.2 & 35.3 & 68 & 97.3 & 71.56 \\
			& AdaBN   & 95.2 & 98.6 & 98.7 & 72.2 & 98.6 & 98.9 & 97.2 & 96.7 & 91.1 & 64.6 & 91.8 & 30.1 & 84.8 & 88.9 & 85.6 & 86.2\\ 
			& SHOT-IM & 98.6 & \textbf{99.2} & \textbf{99} & 98.1 & \textbf{99.2} & 99.1 & 99 & 98.2 & \textbf{99.2} & 75.1 & \textbf{98.8} & \textbf{99.3} & 99.3 & \textbf{98.1} & \textbf{99.1} & 97.3\\ 
			& FR & 97.8 & 98.3 & 98.4 & 97.8 & 98.6 & 98.3 & 98 & 97.5 & 98.3 & 76.7 & 98.2 & 98.6 & 98.7 & 97.2 & 98.1 & 96.7\\
			& BUFR & \textbf{98.7} & 99 & 98.8 & \textbf{98.4} & 99.1 & 98.8 & 98.7 & 97.9 & 99 & 64.5 & 98.7 & 99.2 & 99.1 & 97.9 & 98.7 & 94.4\\
			\cmidrule(l){2-18} 
			& $DAFR^2$    & 94.8 & 98.5 & 97.7 & 96.2 & 98.8 & \textbf{99.2} & \textbf{99.3} & \textbf{98.3} & 98.1 & \textbf{99.4} & 90.7 & 97.2 & \textbf{99.5} & 94.6 & 98.9 & \textbf{97.41} \\ 
			
			\bottomrule
		\end{tabular}%
	}
	\label{tab:tea_comparison_large_margins}
\end{table}

\newpage
\section{Further analysis results}
This appendix presents the results of two additional experiments that further demonstrate the effectiveness of $DAFR^2$ in domain adaptation. First, we analyze the feature space distance for misclassified data points. Specifically, we compare the distance between the feature mapping of data points incorrectly classified by a baseline model (without $DAFR^2$) and their closest same-class feature mapping, to the same distance when $DAFR^2$ is applied. This helps illustrate how $DAFR^2$ affects the clustering of features in the embedding space. Second, we investigate the cross-entropy (CE) loss for data points initially misclassified by a baseline model. We compare the CE loss for these data points before and after $DAFR^2$ is applied, revealing interesting information on how $DAFR^2$ reduces classification errors.

These analyses provide further evidence of $DAFR^2$'s ability to improve model performance in domain adaptation tasks. For both experiments, we kept the same experimental setup as described in Appendix B.

\subsection{Impact on feature mapping distances}
\begin{figure}[h!] 
	\centering 
	\includegraphics[width=0.7\textwidth]{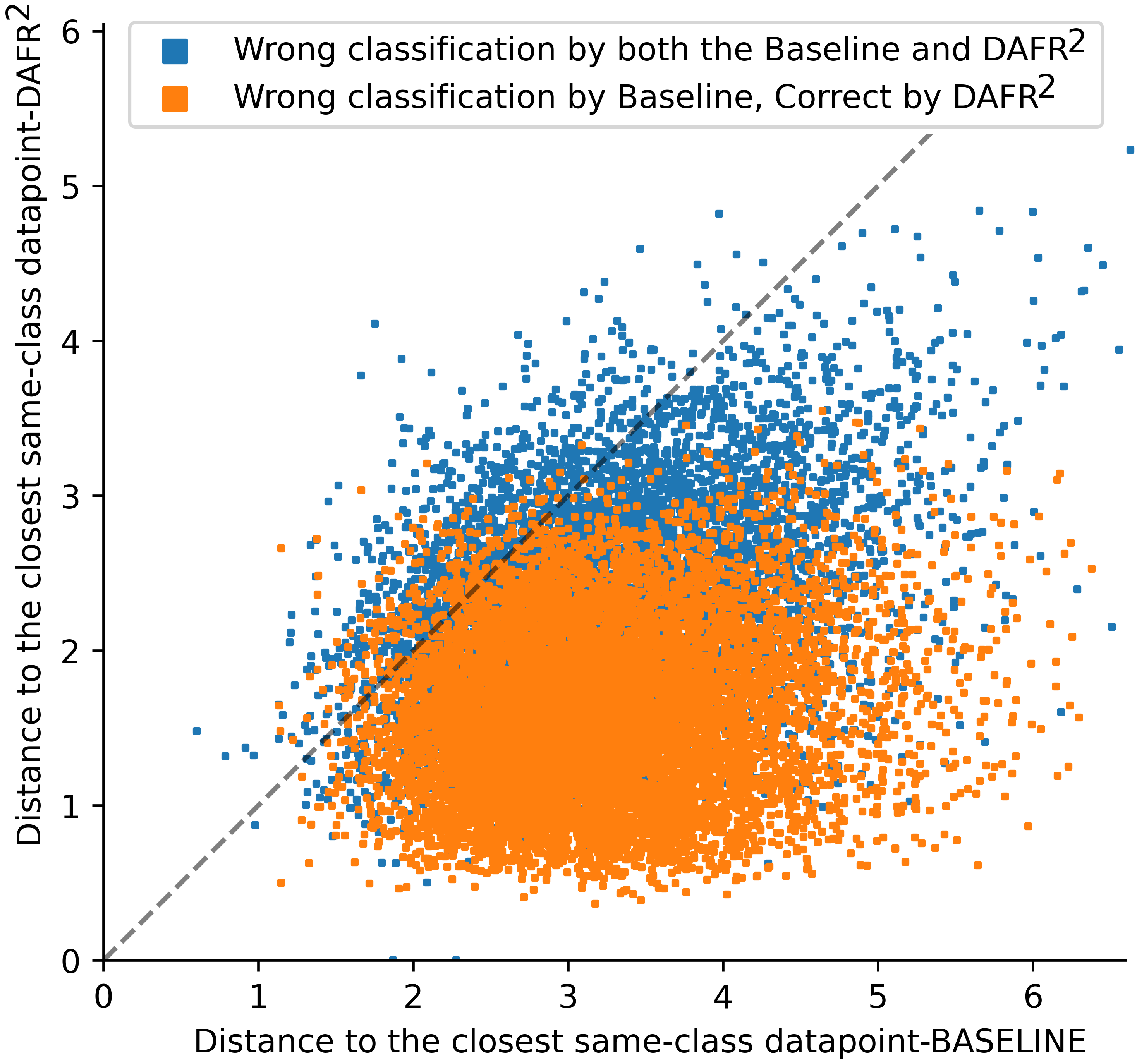} 
	\caption{We compare the Euclidean distances of feature mappings for data points initially misclassified by the baseline model. The x-axis represents these distances for the baseline model and the y-axis shows the distances when $DAFR^2$ is employed. This comparison clearly demonstrates that our proposed algorithm significantly reduces misclassifications and consistently brings data points closer to their correct class feature space. (Best viewed in color.)}
	\label{fig:featureDistanceAnalysis} 
\end{figure}

This analysis focuses on the data points within the CIFAR10-C test set that the baseline model trained exclusively on the CIFAR-10 dataset incorrectly classifies. We aimed to compare the Euclidean distance between the feature mapping of each of these misclassified data points and the feature mapping of its nearest neighbor within the same class. This comparison was carried out for two scenarios: with and without the application of $DAFR^2$. The results, presented in Figure \ref{fig:featureDistanceAnalysis}, demonstrate that our algorithm successfully corrects the classification of most data points that the baseline model misidentified, indicated by the orange coloring in the figure. The vast majority of points, especially those whose classification is corrected by $DAFR^2$, lies below the 45-degree dotted line. This indicates a significant reduction in the distance between the feature mappings of these data points and their correct class feature space, suggesting effective feature alignment between the source and the target models.  

\subsection{Impact on CE loss}
\begin{figure}[h!] 
	\centering 
	\includegraphics[width=0.7\textwidth]{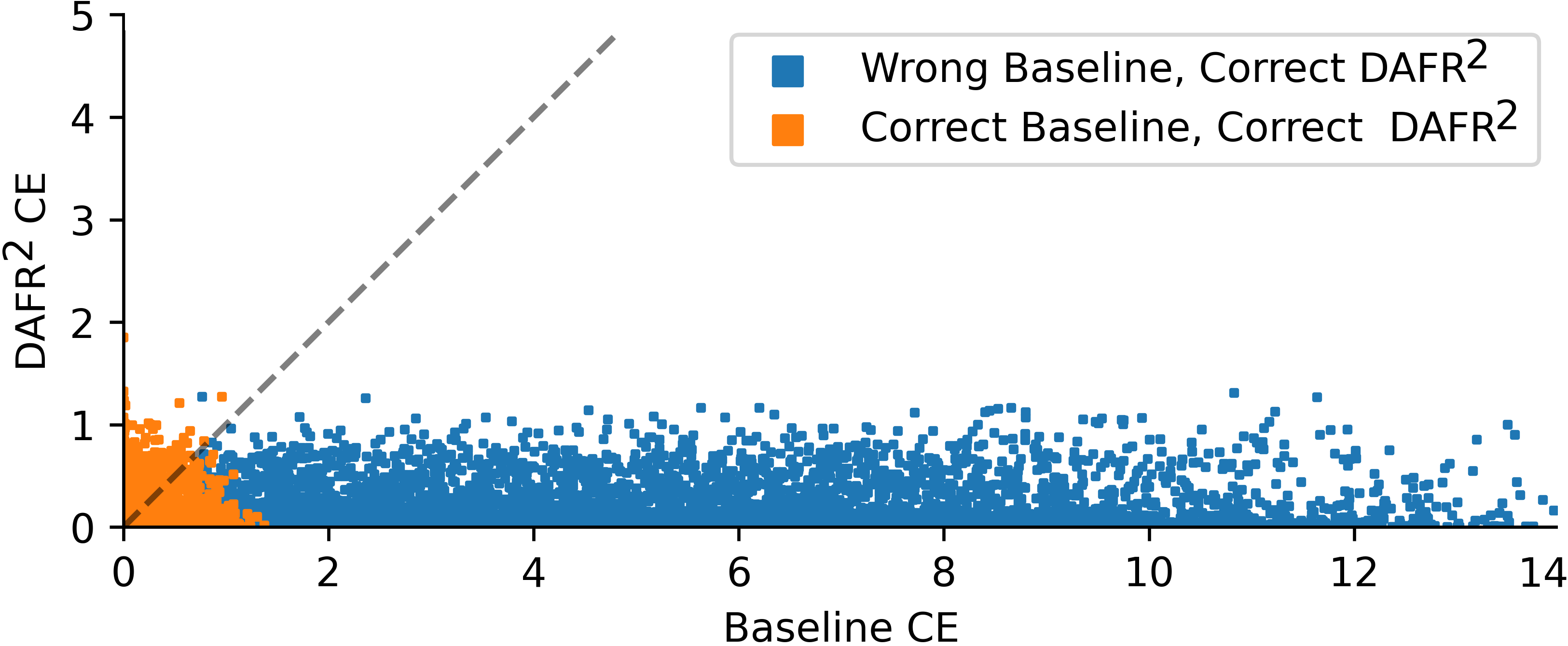} 
	\caption{We compare the CE loss of data points from the CIFAR10-C test set that $DAFR^2$ correctly classifies. For points also correctly classified by the baseline model, both models exhibit low CE losses. However, data points that the baseline model incorrectly classifies show significantly lower CE loss when processed by our algorithm, as indicated by low y-axis and high x-axis values in the plot. Best viewed in color.}
	\label{fig:CEAnalysis} 
\end{figure}

This analysis focuses on how $DAFR^2$ impacts the CE loss of the data points in the CIFAR10-C test set it correctly classifies, regardless of whether the baseline model (trained solely on CIFAR-10) classified them correctly or incorrectly. Figure \ref{fig:CEAnalysis} presents the results of this analysis. The CE loss for data points correctly classified by both the baseline and $DAFR^2$ models doesn't reveal any notable insights. However, for data points where $DAFR^2$ corrects the classification, we observe a significant reduction in their CE loss with the vast majority of the points lying below the 45-degree dotted line. This suggests that domain adaptation with $DAFR^2$ produces a feature space on which the principal structures that the two domains share are properly aligned.

\end{document}